\documentclass{article} % For LaTeX2e
\usepackage[table]{xcolor}
\usepackage{iclr2025_conference}
\usepackage{times}
\usepackage{amsmath,amsfonts,bm}

\def\eqref#1{equation~\ref{#1}}
\def\1{\bm{1}}

\DeclareMathAlphabet{\mathsfit}{\encodingdefault}{\sfdefault}{m}{sl}
\SetMathAlphabet{\mathsfit}{bold}{\encodingdefault}{\sfdefault}{bx}{n}

\usepackage{caption}
\usepackage{algorithm}
\usepackage{algorithmic}
\usepackage{booktabs}
\usepackage{multirow}
\usepackage{array}
\usepackage{overpic}
\usepackage{hyperref}
\usepackage{url}
\usepackage{makecell}
\usepackage{enumitem}
\usepackage{graphicx}
\usepackage{subcaption}
\usepackage{wrapfig}
\usepackage{setspace}

\hypersetup{
    colorlinks=true,
    linkcolor=red,
    citecolor=blue,
    urlcolor=red
}

\makeatletter
\renewcommand{\thanks}[1]{%
  \protected@xdef\@thanks{\@thanks\protect\footnotetext[0]{#1}}%
}
\makeatother

\definecolor{ao(english)}{rgb}{0.0, 0.5, 0.0}
\newcommand{\revise}[1]{#1}
\newcommand{\name}{ChartMoE}
\newcommand{\alignname}{ChartMoE-Align}

\title{ChartMoE: Mixture of Diversely Aligned Expert Connector for Chart Understanding}

\author{\hspace{-5pt}Zhengzhuo Xu$^{12*}$\thanks{\noindent $^*$Equal contributions. $^{\dagger}$Corresponding authors.} Bowen Qu$^{13*}$ Yiyan Qi$^{1*}$ Sinan Du$^{2}$ Chengjin Xu$^{1}$ Chun Yuan$^{2\dagger}$ Jian Guo$^{14\dagger}$ \\
\hspace{25pt}\textsuperscript{1}International Digital Economy Academy \ \textsuperscript{2}Tsinghua University \  \textsuperscript{3}Peking University \\
\hspace{55pt}\textsuperscript{4}Hong Kong University of Science and Technology (Guangzhou)\\
\hspace{75pt}\url{https://github.com/IDEA-FinAI/ChartMoE}
}

\iclrfinalcopy % Uncomment for camera-ready version, but NOT for submission.
\begin{document}

\maketitle
\vspace{-15pt}
\begin{abstract}
Automatic chart understanding is crucial for content comprehension and document parsing. Multimodal Large Language Models (MLLMs) have demonstrated remarkable capabilities in chart understanding through domain-specific alignment and fine-tuning. However, current MLLMs still struggle to provide faithful data and reliable analysis only based on charts. To address it, we propose ChartMoE, which employs the Mixture of Expert (MoE) architecture to replace the traditional linear projector to bridge the modality gap. Specifically, we train several linear connectors through distinct alignment tasks, which are utilized as the foundational initialization parameters for different experts. Additionally, we introduce ChartMoE-Align, a dataset with nearly 1 million chart-table-JSON-code quadruples to conduct three alignment tasks (chart-table/JSON/code). Combined with the vanilla connector, we initialize different experts diversely and adopt high-quality knowledge learning to further refine the MoE connector and LLM parameters. Extensive experiments demonstrate the effectiveness of the MoE connector and our initialization strategy, e.g., ChartMoE improves the accuracy of the previous state-of-the-art from 80.48\% to 84.64\% on the ChartQA benchmark.
\end{abstract}
\vspace{-10pt}

\section{Introduction}
\vspace{-5pt}

Charts serve as a fundamental tool for data visualization, with automated chart interpretation gaining prominence in domains such as text analysis~\cite{introtext}, scientific research~\cite{SciCap}, and policy-making~\cite{chartinsights}. Chart understanding is a complex task that demands the identification of visual cues, the comprehension of intricate interactions, and the precise inference of values informed by prior knowledge~\cite{awesome-chart}. Previous work~\cite{MatCha, DePlot} typically pre-trained on domain-specific charts, which are constrained by limited resources and narrow task focus. In contrast, Multi-modal Large Language Models (MLLMs)~\cite{BLIP2, llava, qwen, Mplug, Minigptv2, GPT4} exhibit substantial potential in image comprehension and instruction following. The community has achieved advanced progress by creating chart understanding datasets~\cite{MMC, Chartllama, chartgemma, ChartBench} and applying supervised fine-tuning based on well-performed MLLMs~\cite{ChartAst, ChartReformer}. With the exponential growth of chart data, automated chart interpretation via MLLMs is emerging as a promising avenue.

Recent studies advocate for chart alignment as a foundational step for LLaVA-like MLLMs~\cite{llava, internlm-xcomposer, BLIP3}, which bridge the visual encoder and LLM through MLP connector. They usually utilize chart-to-table alignment task to train the connector effectively~\cite{ChartAst, ChartReformer, DocOwl}. However, tables only provide basic information, such as numerical values and titles, which fail to capture the full range of chart elements. Despite some efforts to align with more informative text~\cite{ChartReformer}, the heavy alignment tasks may lead to the erosion of the connector's general capabilities, e.g., instruction following and visual counting, which are derived from the pre-training on large-scale visual-language data. To mitigate knowledge forgetting, one intuitive approach is to further tune with its original data, which results in redundant training and computational burden. 

\begin{figure}[t]
    \centering
    \includegraphics[width=\textwidth]{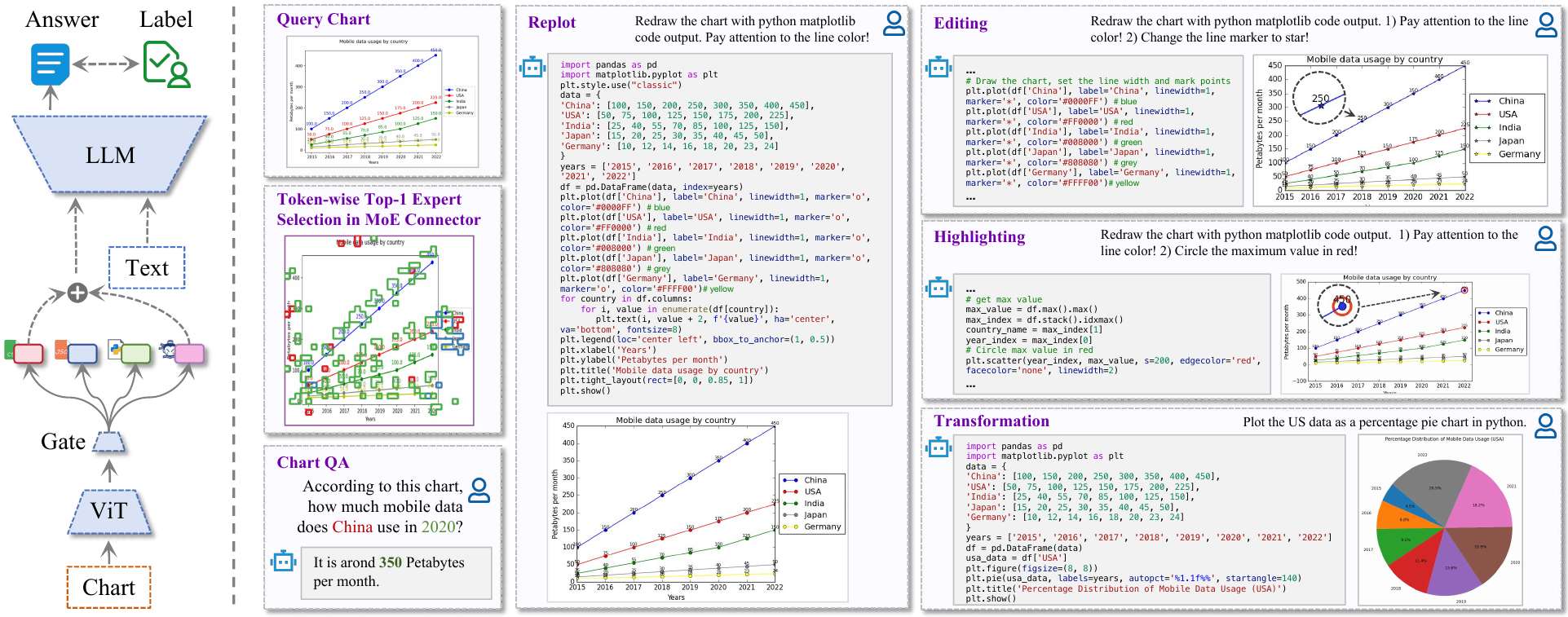}
    \captionsetup{skip=5pt}
    \caption{Overview and capabilities of \name: We introduce a MoE architecture connector and provide visualizations of the top-1 expert selection (refer to Fig.~\ref{fig_token_wise_expert} and Appendix~\ref{apdx_sec_visual_token} for details). \name\ can extract highly precise values and provide flexible chart editing through code-based interactions.}
    \vspace{-17pt}
    \label{fig_teaser}
\end{figure}

In this paper, we try to address these challenges via Mixture of Experts (MoE) architecture~\cite{stmoe}. MoE enhances model capacity by activating a subset of experts through a router. Since the alignment tasks work on the connector, we replace only the MLP projector with MoE while keeping the vision encoder and LLM frozen. Our insight lies in the \textbf{\textit{expert initialization manner}}. Random initialization can lead to training instability and convergence at sub-optimal points (Fig.~\ref{fig_loss}). Recent co-upcycling initialization~\cite{CoUpcycling} addresses this issue by duplicating the vanilla connector parameters across all experts. However, it fails to avoid the dilemma of expert homogenization, where the experts end up with similar functionalities. 

In contrast, we attempt to inject distinct prior knowledge into each expert first to tackle these challenges. Unlike natural images, \textit{charts can be represented in various text formats, e.g., tables, attribute JSON, and rendering code}. As shown in Fig.~\ref{fig_teaser}\&~\ref{fig_train_pipeline}, in addition to chart-table, we introduce chart-JSON alignment to capture detailed elements like color or topological relationships and chart-code alignment to incorporate rendering details such as numerical values, color hex codes, and visual elements interactions (refer to Appendix~\ref{apdx_sec_chartmoealign}). We independently conduct various alignment tasks to capture more diverse chart features and thus obtain three distinct initialization approaches. We also retain the vanilla connector to preserve the capabilities of the MLLM on general tasks effectively. 

Building upon the proposed four initialization manners, we introduce \name, an SFT-based MLLM with MoE connector for chart comprehension and reasoning. Our purpose in applying MoE connector is not to increase model capacity, but rather to improve chart comprehension and diverse representation through alignment tasks, while maintaining performance on other general tasks. Hence, we preserve the original connector parameters as one expert initialization manner. The MoE connector is extremely lightweight, so it adds negligible computational cost during both training and inference. Interestingly, we observe that experts in ChartMoE exhibit distinct visual token preferences, e.g., the vanilla expert favors background tokens while other experts focus more on tokens with legends or numbers (Fig.~\ref{fig_token_wise_expert} and Appendix~\ref{apdx_sec_visual_token}). Considering that the distribution of visual tokens is naturally imbalanced in chart scenarios, we remove the expert-balanced loss in MoE and obtain further performance gain. Due to the scarcity of rich structural text for chart alignment, we design a pipeline (Fig.~\ref{fig_align_pipeline}) to generate nearly 1 million quadruplets chart-table-JSON-code to build the \alignname\ dataset for alignment. We train \name\ in 3 stages. First, we initialize experts via the proposed four manners. Then, we conduct high-quality knowledge learning using the MMC instruction~\cite{MMC} to train the routing network, expert connectors, and LoRA~\cite{lora} modules. Finally, we employ annealing training on ChartQA~\cite{ChartQA} and ChartGemma~\cite{chartgemma}. \name\ achieves state-of-the-art (SOTA) performance and provides more precise numbers and comprehensive attributes of charts (Appendix~\ref{apdx_sec_chatdemo}). Refer to Appendix~\ref{apdx_contribution} for detailed comparisons with other MoE works w.r.t. \textit{motivation}, \textit{initialization}, and \textit{complexity}.  In summary, our contributions are:

\setlength{\leftmargini}{15pt}
\begin{enumerate}[itemsep=0pt, topsep=0pt, label=\alph*)]
    \item We present \name\ for faithful and reasonable chart understanding, with the connector based on Mixture of Expert architecture, to bridge the chart and LLM branches. All experts are initialized based on various alignment training tasks to avoid expert homogenization.
    \item We introduce \alignname, a large-scale dataset with nearly 1 million meticulous chart-table-JSON-code quadruplets for chart alignment pre-training.
    \item We propose to train \name\ with a three-stage training paradigm, including connector alignment pre-training, high-quality knowledge learning, and annealing chart tuning.
    \item Extensive quantitative and qualitative studies demonstrate that \name\ significantly outperforms previous state-of-the-art across several benchmarks by a large margin.
\end{enumerate}

\section{Related Work}

\textbf{Multimodal large language models} leverages a connector to bridge the gap between large language models~\cite{llama, GPT-1, GPT-3, OPT, vicuna} and vision encoders~\cite{clip, dinov2} to enable enriched capabilities of comprehension and instruction following. Approaches such as BLIP2~\cite{BLIP2}, Flamingo~\cite{Flamingo}, mPLUG-Owl~\cite{Mplug}, and Qwen-VL~\cite{qwen-vl} utilize QFormers or Resamplers to align modalities on extensive datasets of image-text pairs. LLaVA~\cite{llava, Improvedllava} is the pioneering work to extend the instruction tuning paradigm to visual tasks with text-only GPT-4~\cite{GPT4}, achieving tremendous performance using a simple MLP without compromising visual information to refine the multimodal alignment. Some works~\cite{sphinx, MMVP-MoF, cambrian-1} explore the combination of various vision encoders, complementarily enhancing visual representations to bolster the fine-grained visual perception of MLLMs. Despite efforts in structural design, training strategies and data quality remain crucial in the advancement of MLLMs.

\textbf{Chart Reasoning} refers to chart analysis, summarization, and etc. Existing methods can be categorized as 1) \textit{Two-stage methods} use specialized extraction modules to generate intermediary representations of chart information, like tables, which are provided as textual prompts for LLMs. Pix2Struct~\cite{Pix2Str} aligns markdown data with charts. MatCha~\cite{MatCha} aligns various data formats (e.g., tables and code) with charts on several downstream tasks. DePlot~\cite{DePlot} fine-tunes Pix2Struct for table extraction and uses LLMs to process queries based on the extracted data. ChartVLM~\cite{ChartVLM} employs a discriminator to ascertain the necessity of intervention by LLMs for a given query. 2) \textit{End-to-end methods} strive to tackle chart reasoning challenges with a unified model. ChartLlama~\cite{Chartllama} incorporates diverse charts and downstream tasks based on LLaVA~\cite{llava}. ChartPaLI~\cite{ChartPaLI}, ChartAst~\cite{ChartAst}, and MMC~\cite{MMC} conduct alignment on table-chart pairs. UReader~\cite{Ureader} aligns all data with markdown, while mPLUG-Owl2~\cite{mPLUG2} achieves superior performance with high-resolution inputs. ChartThinker~\cite{ChartThinker} and DOMINO~\cite{DOMINO} propose the CoT~\cite{CoT} for chart reasoning. LaMenDa~\cite{LAMENDA} trains MLLMs via step-by-step reasoning QA. ChartReformer~\cite{ChartReformer} introduces chart-JSON alignment, while OneChart~\cite{OneChart} aligns charts with Python dictionaries. MiniGPT-v2~\cite{Minigptv2}, Doc-Owl~\cite{DocOwl}, and TinyChart~\cite{TinyChart} tackle the reasoning efficiency for high-resolution charts by merging tokens.

\section{\name}

\subsection{Architecture}
\label{sec_method_1}

The \name\ is based on InternlmXC-v2~\cite{internlm-xcomposerv2} due to the concise LLaVA-like architecture~\cite{llava} and performance on par with GPT-4 on text-image comprehension. The base model includes a vision encoder and a LLM connected by a two-layer MLP. \name\ replaces the MLP with a MoE architecture as the connector to leverage diverse prior knowledge. 

\textbf{Vision Encoder.}
We utilize CLIP ViT-Large~\cite{clip} as the vision encoder, leveraging its rich prior knowledge gained from training on millions of image-text pairs. Considering the impact of chart resolution on performance, we set the input resolution to 490 $\times$ 490 to strike a balance between efficiency and performance. Formally, the visual encoder $\mathcal{M}^{V}(\cdot)$ will project the chart $\mathcal{I}$ into $N$ tokens $V:=\{v_1, v_2, \ldots, v_N\}$, where $N=1225$ in the \name.

\begin{figure*}[t!]
    % \hsize=\textwidth
    \centering
      \begin{overpic}[width=\linewidth, grid=False]{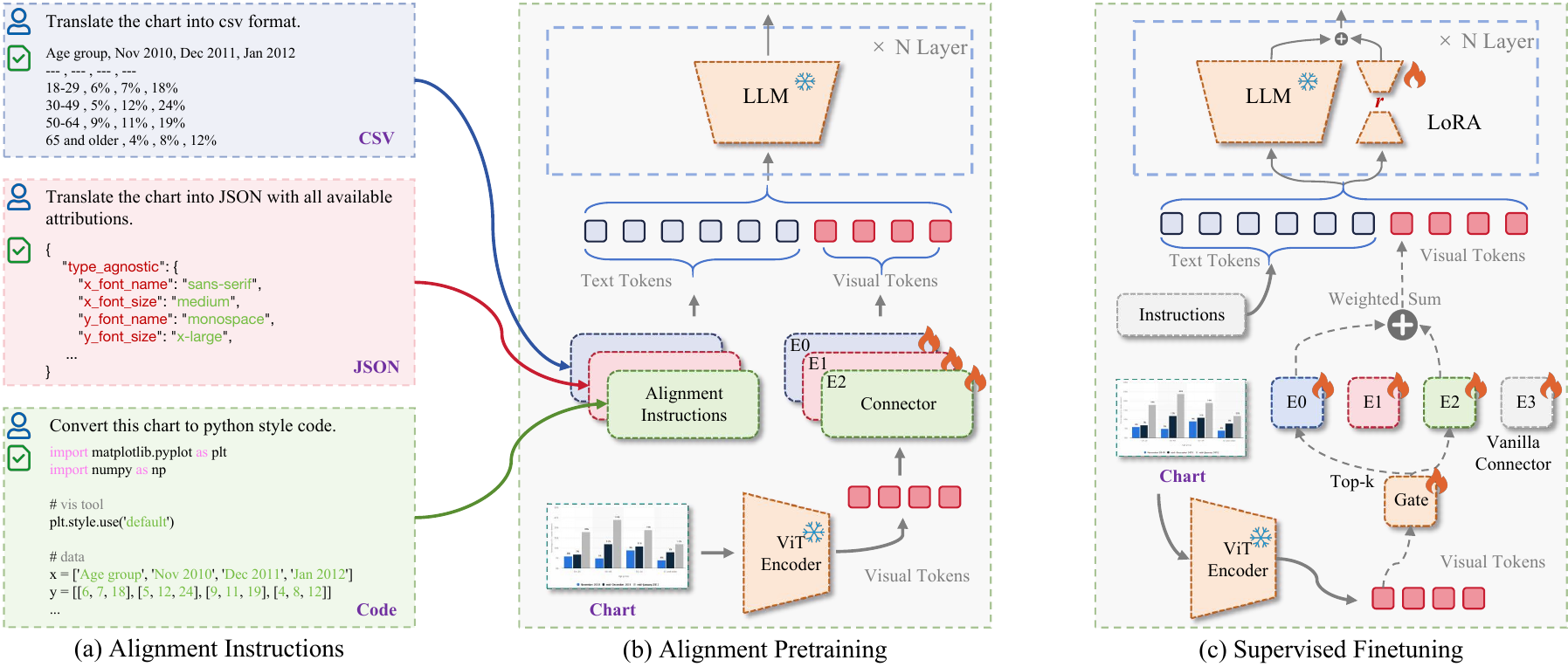}
      \end{overpic}
      \captionsetup{skip=5pt}
      \caption{Overview of proposed \name. (a) Examples of alignment instructions. (b) We conduct three different alignment tasks in parallel. (c) We initialize MoE connectors in four different manners and train the gate network, experts, and LoRA during the supervised fine-tuning stage.}
      % \vspace{-20pt}
    \label{fig_train_pipeline}
\end{figure*}

\textbf{Mixture-of-Experts Connector.}
As illustrated in Fig.~\ref{fig_train_pipeline}c, the MoE architecture employs a parallel multi-expert collaboration approach. This architecture comprises $L$ experts $\mathcal{M}^E(\cdot)$, each designed with the same linear layer as the baseline. For a visual token $v_i$ given by $\mathcal{M}^V$, the gating network $\mathcal{M}^G(\cdot)$ will calculate the routing weight $g_j(v_i)$ of each expert $\mathcal{M}^E_j(\cdot)$ and select top-\textit{K} to activate. Finally, the tokens processed by each expert $\mathcal{M}^E_j$ will be averaged according to the weight $g_j(v_i)$ given by $\mathcal{M}^G$ to get the token $\hat{v}_i$ for the LLM branch $\mathcal{M}^L$.

\textbf{Large Language Model.}
Following the baseline, we employ the \textit{InternLM2-7B-ChatSFT} variant as the LLM $\mathcal{M}^L$, implemented as a transformer decoder with a causal attention mask.
We concate the visual tokens $\hat{V}:=\{\hat{v}_1, \hat{v}_2, \ldots, \hat{v}_N\}$ given by MoE connector with the $M$ input text $\mathcal{T}$ tokens $T:=\{t_1, t_2, \ldots, t_M\}$ to form the input token sequence for the LLM $\mathcal{M}^L$. Formally, given the chart $\mathcal{I}$ and instruction $\mathcal{T}$, the output $\mathcal{O}$ of proposed \name\ can be formulated as:
\begin{equation}
    \{v_1, v_2, \ldots, v_N\} = \mathcal{M}^{V}(\mathcal{I}),
\end{equation}
\begin{equation}
    \hat{v_i} = \sum_j^L g_j(v_i) \mathcal{M}^E_j(v_i), \quad  g(v_i) = \mathrm{Top} (\sigma(\mathcal{M}^G(v_i)); K),
\end{equation}
\begin{equation}
    \mathcal{O} = \mathcal{M}^L(\{\hat{v}_1, \hat{v}_2, \ldots, \hat{v}_N; t_1, t_2, \ldots, t_M\}),
\end{equation}
where $\sigma$ indicates \textit{softmax} and the $\mathrm{Top}(\cdot;K)$ will reset the non-Top K routing weight to 0.

\textbf{Initialization of Expert.}
Previous approaches initialize expert parameters via \textit{1) Random initialization}, which may lead to convergence difficulties during supervised fine-tuning, and \textit{2) Co-upcycling initialization}~\cite{CoUpcycling}, i.e., copy baseline connector parameters to each expert, which may lead to homogenization of experts. \name\ proposes initializing experts' parameters through distinct alignment tasks. We eliminate the load-balancing loss typically used in standard MoE architectures to equalize expert activation frequencies, as our initialization approach allows each expert to specialize in its preferred visual tokens, which inherently exhibit biased distributions.

\subsection{Alignment Pre-training.}
\label{sec_method_2}\vspace{-5pt}
The key insight of \name\ is the experts' initialization parameters from the different alignment pre-training (Fig.~\ref{fig_train_pipeline}a). Specifically, as illustrated in Fig.~\ref{fig_train_pipeline}b, we align expert connectors using three distinct alignment tasks, where only the connector parameters will be updated. We visualize the \textit{visual token preferences} of each expert for both chart (Fig.~\ref{fig_token_wise_expert}\&~\ref{fig_apdx_tokenwise_chart}) and non-chart (Fig.~\ref{fig_apdx_tokenwise_llava}) images.

\textbf{Alignment with Table.} 
Charts convey key information that can be more precisely expressed in tabular form, and LLMs are particularly adept at processing such structured data. Hence, we introduce a chart-table alignment task, aiming to translate chart content into tabular format. The connector is trained to convert chart information into corresponding CSV tables, thereby improving model performance in numerical data extraction and chart interpretation. 

\textbf{Alignment with JSON.}
Although tables capture the numerical information from charts, they miss semantic details such as colors, shapes, and fonts. To fill this gap, we propose a chart-JSON alignment task, which represents chart attributes in JSON format. This task requires the connector to focus not only on the numerical data but also on visual and semantic properties. Accurately extracting chart attributes is essential for tasks like chart redrawing and editing.

\vspace{-5pt}
\begin{wraptable}{r}{0.45\textwidth}
\centering
\captionsetup{skip=0pt}
\caption{Datasets used for training \name. We conduct alignment pre-training with synthetic data and supervised tuning with high-quality, real-world data. Only ChartQA is used in the ablation due to GPU constraints.}
\resizebox{\linewidth}{!}{
\setlength{\tabcolsep}{2pt}
\begin{tabular}{@{}cccc@{}}
\toprule[1.5pt]
\multicolumn{1}{c|}{Source} & \multicolumn{1}{c|}{Data Type} & \multicolumn{1}{c|}{Task Type} & \multicolumn{1}{c}{Samples} \\ \midrule[0.5pt]
\multicolumn{4}{c}{\textit{Alignment Training}} \\
\multicolumn{1}{c|}{\multirow{3}{*}{ChartQA}} & \multicolumn{1}{c|}{\multirow{3}{*}{synthetic}} & \multicolumn{1}{c|}{\texttt{chart} to \texttt{table}} & 18.3K \\
\multicolumn{1}{c|}{} & \multicolumn{1}{c|}{} & \multicolumn{1}{c|}{\texttt{chart} to \texttt{JSON}} & 18.3K \\
\multicolumn{1}{c|}{} & \multicolumn{1}{c|}{} & \multicolumn{1}{c|}{\texttt{chart} to \texttt{code}} & 18.3K \\ \midrule
\multicolumn{1}{c|}{\multirow{3}{*}{PlotQA}} & \multicolumn{1}{c|}{\multirow{3}{*}{synthetic}} & \multicolumn{1}{c|}{\texttt{chart} to \texttt{table}} & 157K \\
\multicolumn{1}{c|}{} & \multicolumn{1}{c|}{} & \multicolumn{1}{c|}{\texttt{chart} to \texttt{JSON}} & 157K \\
\multicolumn{1}{c|}{} & \multicolumn{1}{c|}{} & \multicolumn{1}{c|}{\texttt{chart} to \texttt{code}} & 157K \\ \midrule
\multicolumn{1}{c|}{\multirow{3}{*}{ChartY}} & \multicolumn{1}{c|}{\multirow{3}{*}{synthetic}} & \multicolumn{1}{c|}{\texttt{chart} to \texttt{table}} & 763.6K \\
\multicolumn{1}{c|}{} & \multicolumn{1}{c|}{} & \multicolumn{1}{c|}{\texttt{chart} to \texttt{JSON}} & 763.6K \\
\multicolumn{1}{c|}{} & \multicolumn{1}{c|}{} & \multicolumn{1}{c|}{\texttt{chart} to \texttt{code}} & 763.6K \\ \midrule
\multicolumn{3}{l|}{Total} & 2.8M \\
\multicolumn{3}{l|}{Usage: \texttt{Table} = 500K \texttt{JSON} = 200K \texttt{Code} = 100K} & 800K \\ \midrule
\multicolumn{4}{c}{\textit{High-Quality Knowledge Learning}} \\
\multicolumn{1}{c|}{\multirow{2}{*}{MMC}} & \multicolumn{1}{c|}{synthetic} & \multicolumn{1}{c|}{QA \& reasoning} & \multirow{2}{*}{410K} \\
\multicolumn{1}{c|}{} & \multicolumn{1}{c|}{\& real world} & \multicolumn{1}{c|}{\& summariztion} &  \\ \midrule
\multicolumn{4}{c}{\textit{Chart Specific Annealing Tuning}} \\
\multicolumn{1}{c|}{ChartQA} & \multicolumn{1}{c|}{real world} & \multicolumn{1}{c|}{QA} & 28.3K$\times$2 \\
\multicolumn{1}{c|}{\multirow{2}{*}{ChartGemma}} & \multicolumn{1}{c|}{\multirow{2}{*}{real world}} & \multicolumn{1}{c|}{QA \& PoT \& reasoning} & \multirow{2}{*}{163.2K} \\
\multicolumn{1}{c|}{} & \multicolumn{1}{c|}{} & \multicolumn{1}{c|}{\& summariztion} &  \\ \midrule
\multicolumn{3}{l|}{Total} & 220.8K \\ 
\bottomrule[1.5pt]
\end{tabular}
}
\vspace{-15pt}
\label{tab_dataset}
% \end{table}
\end{wraptable}

\textbf{Alignment with Code.}
To fully align with charts, we further introduce a chart-code alignment task. Since the underlying drawing code fully defines a chart, this approach enables the connector to convert the chart's visual tokens into representations in the LLM domain. Notably, we provide the drawing code explicitly, including precise numerical values and rendering attributes, e.g., numbers represented in Python lists and colors in hexadecimal code. Refer to Fig.~\ref{fig_example_of_chartalign} for more detailed cases. The code enables the model to perform in-depth summarization, analysis, and editing of charts. This expert is significantly more sensitive to the trends and key elements in the charts.

\begin{figure*}[t!]
    % \hsize=\textwidth
    \centering
      \begin{overpic}[width=1\linewidth, grid=False]{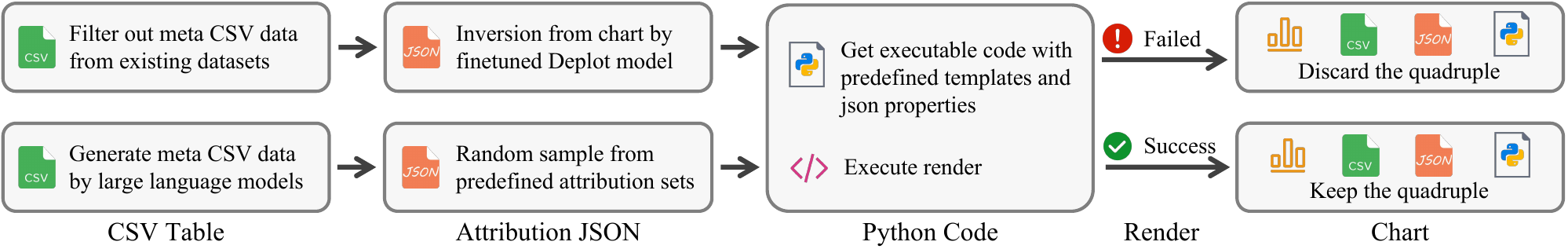}
      \end{overpic}
      \captionsetup{skip=3pt}      
      \caption{Overview of \alignname\ data generation pipeline. The charts are plotted by Python \textit{matplotlib}.}
      \vspace{-20pt}
    \label{fig_align_pipeline}
\end{figure*}

\textbf{\alignname \  generation pipeline.} As Fig.~\ref{fig_align_pipeline} illustrates, 1) We filter charts with meta CSV from existing datasets~\cite{ChartQA, PlotQA} and data generated by LLMs~\cite{OneChart}. 2) We use a fine-tuned Deplot~\cite{DePlot} to inverse the plotting attributes following the templates provided by ChartReformer and randomly sample missing attributes from the predefined set. 3) We create code templates for different types of charts and generate plotting code based on the meta CSV and extracted JSON attributes. \textit{Note that all values and attributes in the code are explicitly represented.} 4) We retain the \texttt{(table, JSON, code, chart)} quadruples that pass compilation. \revise{Tab.~\ref{tab_dataset} shows the data sources \& size and refer to Appendix~\ref{apdx_sec_chartmoealign} for details.}

\subsection{Supervised Fine-tuning.}
\label{sec_method_3}

We initialize \name\ using the structure shown in Fig.~\ref{fig_train_pipeline}c after aligning the connectors across 3 distinct tasks separately. We also retain the vanilla connector to maintain the baseline's excellent dialogue capabilities, which aligns with the principle of residual optimization~\cite{resnet}. We train the MoE connector and LLM during this stage with LoRA~\cite{lora}, as shown in Fig.~\ref{fig_train_pipeline}c. Considering the training principles proposed in LLaVA-NeXT~\cite{llavanext}, this stage is divided into high-quality knowledge learning and chart-specific annealing training.

\textbf{High-Quality Knowledge Learning.}
We adopt MMC~\cite{MMC} to enhance the \name's knowledge. MMC includes a variety of chart types and tasks such as chart-related question answering, translation, extraction, reasoning, and analysis. Considering data quality, we only utilize the MMC-Instruction subset, which has been manually verified. Notice that the quality of instruction data is more important than quantity in this stage.

\textbf{Chart Specific Annealing Tuning.}
Following Llama-v3.1~\cite{llama3}, we perform annealing tuning before evaluating mainstream benchmarks. We increase the learning rate and conduct instruction tuning using the training sets of ChartQA and ChartGemma to adjust the query styles and answer formats of these benchmarks.

\textbf{Program of Thought (PoT) Inference.} We require the model to generate the variables and operation code rather than producing direct answers. This inference pipeline addresses the mathematical capabilities by employing \textit{Python} to handle the logical computations, which is the shortcoming of all open-sourced models. With better numerical extraction abilities, PoT can significantly enhance our \name's question-answering performance.

\section{Experiment}

\subsection{Implementation Details}
\vspace{-5pt}

During the alignment stage, we train the connector parameters and keep the visual encoder and LLM parameters frozen. In the supervised fine-tuning stage, we continue training the MoE connector while employing LoRA to update the LLM parameters. All training processes are conducted on 4 $\times$ A100-40G GPUs. You can refer to Appendix \ref{apdx_hyper_params} for more details.

\begin{table*}[t]
\centering
\captionsetup{skip=3pt}
\caption{The relaxed accuracy (\%) performance on \textbf{\textit{ChartQA}}. Ada.: Adaptive input resolution. *: Multi-scale image feature, 448$\times$448 in DocOwl. $\dagger$: Employing token merging to reduce computational overhead.}
\resizebox{\linewidth}{!}{
\setlength{\tabcolsep}{5pt}
\begin{tabular}{@{}cccccccccccc@{}}
\toprule[1.5pt]
\multicolumn{1}{c|}{\multirow{2}{*}{Models}} & \multicolumn{1}{c|}{\multirow{2}{*}{Para.}} & \multicolumn{1}{c|}{\multirow{2}{*}{Resolution}} & \multicolumn{3}{c|}{Relax Acc. @0.05} & \multicolumn{3}{c|}{Relax Acc. @0.10} & \multicolumn{3}{c}{Relax Acc. @0.20} \\ \cmidrule(l){4-12} 
\multicolumn{1}{c|}{} & \multicolumn{1}{c|}{} & \multicolumn{1}{c|}{} & Human & Aug. & \multicolumn{1}{c|}{Avg.} & Human & Aug. & \multicolumn{1}{c|}{Avg.} & Human & Aug. & Avg. \\ \midrule
\multicolumn{12}{c}{General MLLMs} \\
\multicolumn{1}{c|}{LLaVA-v1.5} & \multicolumn{1}{c|}{13B} & \multicolumn{1}{c|}{@336} & 25.36 & 18.56 & \multicolumn{1}{c|}{21.96} & 28.56 & 23.52 & \multicolumn{1}{c|}{26.04} & 32.56 & 30.72 & 31.64 \\
\multicolumn{1}{c|}{Qwen-VL} & \multicolumn{1}{c|}{9.6B} & \multicolumn{1}{c|}{@448} & 40.48 & 79.76 & \multicolumn{1}{c|}{60.12} & 43.20 & 82.56 & \multicolumn{1}{c|}{62.88} & 47.52 & 85.76 & 66.64 \\
\multicolumn{1}{c|}{DocOwl-v1.5} & \multicolumn{1}{c|}{8B} & \multicolumn{1}{c|}{@448*} & 47.44 & 91.52 & \multicolumn{1}{c|}{69.48} & 51.92 & 92.08 & \multicolumn{1}{c|}{72.00} & 56.72 & 93.12 & 74.92 \\
\multicolumn{1}{c|}{InternlmXC-v2} & \multicolumn{1}{c|}{8B} & \multicolumn{1}{c|}{@490} & 62.72 & 81.28 & \multicolumn{1}{c|}{72.00} & 66.72 & 84.08 & \multicolumn{1}{c|}{75.40} & 70.80 & 86.56 & 78.68 \\ \midrule
\multicolumn{12}{c}{Specialist Chart Models} \\
\multicolumn{1}{c|}{Pix2Struct} & \multicolumn{1}{c|}{282M} & \multicolumn{1}{c|}{Ada.} & 30.08 & 76.88 & \multicolumn{1}{c|}{53.48} & 31.68 & 78.40 & \multicolumn{1}{c|}{55.04} & 37.28 & 81.12 & 59.20 \\
\multicolumn{1}{c|}{Matcha} & \multicolumn{1}{c|}{282M} & \multicolumn{1}{c|}{Ada.} & 37.12 & 86.64 & \multicolumn{1}{c|}{61.88} & 39.84 & 87.36 & \multicolumn{1}{c|}{63.60} & 43.52 & 88.56 & 66.04 \\
\multicolumn{1}{c|}{UniChart} & \multicolumn{1}{c|}{201M} & \multicolumn{1}{c|}{@960} & 34.64 & 83.28 & \multicolumn{1}{c|}{58.96} & 36.48 & 84.24 & \multicolumn{1}{c|}{60.36} & 38.88 & 85.28 & 62.08 \\
\multicolumn{1}{c|}{Deplot + LLaVA-v1.6} & \multicolumn{1}{c|}{282M+13B} & \multicolumn{1}{c|}{Ada.} & 53.44 & 87.68 & \multicolumn{1}{c|}{70.56} & 56.80 & 88.48 & \multicolumn{1}{c|}{72.64} & 60.64 & 90.08 & 75.36 \\ \midrule
\multicolumn{12}{c}{Chart MLLMs} \\
\multicolumn{1}{c|}{ChartVLM} & \multicolumn{1}{c|}{13B} & \multicolumn{1}{c|}{Ada.} & 42.08 & 82.48 & \multicolumn{1}{c|}{62.28} & 43.84 & 82.88 & \multicolumn{1}{c|}{63.36} & 46.00 & 83.28 & 64.64 \\
\multicolumn{1}{c|}{OneChart} & \multicolumn{1}{c|}{125M+13B} & \multicolumn{1}{c|}{@1024} & 54.48 & 87.12 & \multicolumn{1}{c|}{70.80} & 57.60 & 87.84 & \multicolumn{1}{c|}{72.72} & 62.00 & 88.64 & 75.32 \\
\multicolumn{1}{c|}{ChartLlama} & \multicolumn{1}{c|}{13B} & \multicolumn{1}{c|}{@336} & 58.40 & \underline{93.12} & \multicolumn{1}{c|}{75.76} & 61.20 & 93.60 & \multicolumn{1}{c|}{77.40} & 63.52 & 94.00 & 78.76 \\
\multicolumn{1}{c|}{ChartGemma+PoT} & \multicolumn{1}{c|}{3B} & \multicolumn{1}{c|}{@448} & 67.84 & 85.28 & \multicolumn{1}{c|}{76.56} & 68.64 & 85.84 & \multicolumn{1}{c|}{77.24} & 69.84 & 86.32 & 78.08 \\
\multicolumn{1}{c|}{TinyChart} & \multicolumn{1}{c|}{3B} & \multicolumn{1}{c|}{@768$\dagger$} & 58.72 & \textbf{94.88} & \multicolumn{1}{c|}{76.80} & 62.56 & \textbf{95.28} & \multicolumn{1}{c|}{78.92} & 67.04 & \textbf{96.16} & 81.60 \\
\multicolumn{1}{c|}{ChartAst} & \multicolumn{1}{c|}{13B} & \multicolumn{1}{c|}{@448} & 64.88 & \underline{93.12} & \multicolumn{1}{c|}{79.00} & 66.24 & \underline{93.84} & \multicolumn{1}{c|}{80.04} & 67.44 & \underline{94.32} & 80.88 \\
\multicolumn{1}{c|}{TinyChart+PoT} & \multicolumn{1}{c|}{3B} & \multicolumn{1}{c|}{@768$\dagger$} & 70.24 & 90.72 & \multicolumn{1}{c|}{80.48} & 71.20 & 91.44 & \multicolumn{1}{c|}{81.32} & 72.40 & 92.56 & 82.48 \\
\rowcolor{blue!5}\multicolumn{1}{c|}{ChartMoE (Ours)} & \multicolumn{1}{c|}{8B} & \multicolumn{1}{c|}{@490} & \underline{71.36} & 91.04 & \multicolumn{1}{c|}{\underline{81.20}} & \underline{75.12} & 92.48 & \multicolumn{1}{c|}{\underline{83.80}} & \underline{78.16} & 93.68 & \underline{85.92} \\
\rowcolor{blue!5}\multicolumn{1}{c|}{ChartMoE+PoT (Ours)} & \multicolumn{1}{c|}{8B} & \multicolumn{1}{c|}{@490} & \textbf{78.32} & 90.96 & \multicolumn{1}{c|}{\textbf{84.64}} & \textbf{80.16} & 92.32 & \multicolumn{1}{c|}{\textbf{86.24}} & \textbf{82.08} & 93.60 & \textbf{87.84} \\ 
\bottomrule[1.5pt]
\end{tabular}
}
\vspace{-20pt}
\label{tab_chartqa}
\end{table*}

\subsection{Evaluation Metrics}
\vspace{-5pt}

\textbf{ChartQA}~\cite{ChartQA} test split consists of 1,250 questions in both human and augmented parts. The charts are three common chart types and are sourced from the real world. It features a variety of human-crafted questions and answers to evaluate models' understanding, reasoning, and data extraction skills. ChartQA adopts relaxed accuracy, which is highlighted shortcomings by recent studies~\cite{OneChart, ChartBench}, such as simplistic string matching and direct float conversion. Therefore, we improve it by 1) using regular expression matching to extract number values, 2) optimizing string matching for short answers, and 3) demonstrating model performance under various relaxed margins. We adopt it for all experiment results.

\textbf{ChartBench}~\cite{ChartBench} focuses on charts without data point annotations. It includes a broader range of chart types, with 9 main categories and 42 subcategories, each containing 50 charts. ChartBench focuses on extracting numerical values, posing a greater challenge as models cannot depend on OCR for precise answers. It adopts \textit{Acc+} for judgments and relaxed accuracy for NQA tasks. The benchmark proposes to extract number values by LLMs first, which is omitted for the stratifying instruction-following ability of \name.

\textbf{ChartFC}~\cite{chartfc} \& \textbf{ChartCheck}~\cite{chartcheck} adopt accuracy to verify whether the claim aligns with the input chart, marking a significant advancement in chart recognition and reasoning abilities. This identifies the potential hallucinations in chart-related contexts. The ChartFC test set has 1,591 questions, and the ChartCheck test set has two splits, containing 937 questions and 981 questions.

\subsection{Comparative Models}
\vspace{-5pt}

\textbf{General MLLMs.}
We compare PaliGemma~\cite{paligemma}, LLaVA-v1.5~\cite{llava} with an MLP connector, Qwen-VL~\cite{qwen-vl} with a Qformer~\cite{BLIP2} connector, DocOwl-v1.5~\cite{DocOwl} that employs multi-level image resolution and token convolution techniques, and the current open-source SOTA, InternlmXC-v2~\cite{internlm-xcomposerv2}.

\textbf{Specialist Chart Models.}
Previous works specifically design models and algorithms for chart question answering. We compare Pix2Struct~\cite{Pix2Str}, Matcha~\cite{MatCha}, UniChart~\cite{unichart}, and Deplot~\cite{DePlot}. Notably, Deplot fails to handle questions in arbitrary formats, so we extract table information with Deplot and use LLaVA-v1.6 to answer the questions.

\textbf{Chart MLLMs.}
Chart-oriented MLLMs are the promising direction for utilizing prior knowledge of LLMs. 
ChartLLaMA~\cite{Chartllama} proposes to generate high-quality instruction data to improve chart question-answering capabilities. 
ChartAst~\cite{ChartAst} suggests aligning the connector with chart-table pairs before supervised fine-tuning. 
ChartVLM~\cite{ChartVLM} uses different decoders to handle different questions based on their difficulty. 
ChartInstruct~\cite{chartinstruct} conducts large-scale chart instruction tuning based on general MLLMs.
OneChart~\cite{OneChart} converts the chart to the table with a dedicated decoder and uses LLMs to answer questions. 
ChartGemma~\cite{chartgemma} proposes more instruction data and achieves efficient chart reasoning based on SigLIP~\cite{SigLIP} and Gemma-2B~\cite{gemma}. 
TinyChart~\cite{TinyChart} adopts token merge to reduce visual tokens and enable high-resolution chart input.

\begin{table*}[t]
\centering
\captionsetup{skip=3pt}
\caption{The zero-shot performance on \textbf{\textit{ChartBench}}. No methods are fine-tuned on the trainset for fairness. We exclude PoT because ChartBench mainly assesses numerical extraction accuracy without math calculation.}
\resizebox{\linewidth}{!}{
\setlength{\tabcolsep}{5pt}
\begin{tabular}{@{}ccccccccccccc@{}}
\toprule[1.5pt]
\multicolumn{1}{c|}{\multirow{2}{*}{Models}} & \multicolumn{4}{c|}{Regular Type} & \multicolumn{7}{c|}{Extra Type} & \multirow{2}{*}{ALL} \\ \cmidrule(lr){2-12}
\multicolumn{1}{c|}{} & Line & Bar & Pie & \multicolumn{1}{c|}{Avg.} & Area & Box & Radar & Scatter & Node & Comb. & \multicolumn{1}{c|}{Avg.} &  \\ \midrule
\multicolumn{13}{c}{General MLLMs} \\
\multicolumn{1}{c|}{LLaVA-v1.5} & 29.12 & 21.26 & 17.28 & \multicolumn{1}{c|}{22.10} & 21.73 & 20.94 & 27.50 & 23.47 & 36.80 & 24.30 & \multicolumn{1}{c|}{24.96} & 23.38 \\
\multicolumn{1}{c|}{Qwen-VL} & 38.00 & 20.71 & 38.24 & \multicolumn{1}{c|}{29.46} & 28.83 & 24.17 & 35.00 & 19.50 & 18.50 & 25.50 & \multicolumn{1}{c|}{26.56} & 28.18 \\
\multicolumn{1}{c|}{DocOwl-v1.5} & 49.60 & 31.69 & 31.54 & \multicolumn{1}{c|}{35.68} & 12.27 & 23.33 & 22.50 & 36.13 & 29.60 & 38.80 & \multicolumn{1}{c|}{27.38} & 32.05 \\
\multicolumn{1}{c|}{Mini-Gemini} & 34.88 & 36.12 & 40.40 & \multicolumn{1}{c|}{36.77} & \underline{31.20} & 23.33 & 30.60 & 35.20 & 43.60 & 27.90 & \multicolumn{1}{c|}{30.61} & 34.37 \\
\multicolumn{1}{c|}{InternlmXC-v2} & \underline{68.16} & \underline{48.74} & \textbf{56.60} & \multicolumn{1}{c|}{\underline{54.50}} & 27.47 & \textbf{25.33} & \underline{40.10} & \underline{52.93} & \underline{50.40} & \underline{46.20} & \multicolumn{1}{c|}{39.72} & \underline{48.41} \\ \midrule
\multicolumn{13}{c}{Specialist Chart Models} \\
\multicolumn{1}{c|}{Pix2Struct} & 2.56 & 2.37 & 1.90 & \multicolumn{1}{c|}{2.33} & 0.13 & 0.13 & 4.60 & 0.67 & 0.40 & 3.20 & \multicolumn{1}{c|}{2.93} & 2.16 \\
\multicolumn{1}{c|}{Matcha} & 6.80 & 5.05 & 3.60 & \multicolumn{1}{c|}{5.18} & 0.27 & 1.60 & 6.20 & 3.46 & 5.40 & 4.80 & \multicolumn{1}{c|}{5.81} & 4.84 \\
\multicolumn{1}{c|}{UniChart} & 7.04 & 5.35 & 4.30 & \multicolumn{1}{c|}{5.55} & 3.86 & 4.80 & 11.60 & 5.06 & 15.80 & 9.60 & \multicolumn{1}{c|}{8.30} & 6.78 \\
\multicolumn{1}{c|}{Deplot+LLaVA-v1.6} & 31.20 & 26.46 & 24.00 & \multicolumn{1}{c|}{27.09} & 21.34 & 13.34 & 24.00 & 41.34 & 42.00 & 31.00 & \multicolumn{1}{c|}{31.57} & 27.62 \\ \midrule
\multicolumn{13}{c}{Chart MLLMs} \\
\multicolumn{1}{c|}{ChartVLM} & 21.92 & 14.16 & 10.50 & \multicolumn{1}{c|}{15.16} & 7.47 & 7.87 & 8.00 & 7.87 & 5.40 & 10.50 & \multicolumn{1}{c|}{8.38} & 11.96 \\
\multicolumn{1}{c|}{ChartLlama} & 26.80 & 18.83 & 20.80 & \multicolumn{1}{c|}{20.99} & 14.27 & 12.00 & 24.30 & 27.73 & 26.20 & 25.80 & \multicolumn{1}{c|}{21.71} & 21.31 \\
\multicolumn{1}{c|}{TinyChart} & 32.40 & 25.81 & 22.50 & \multicolumn{1}{c|}{26.71} & 10.13 & 14.80 & 13.40 & 28.14 & 10.80 & 21.60 & \multicolumn{1}{c|}{22.56} & 22.51 \\
\multicolumn{1}{c|}{OneChart} & 41.28 & 30.28 & 29.60 & \multicolumn{1}{c|}{32.65} & 19.07 & 13.20 & 24.60 & 38.53 & 34.80 & 27.90 & \multicolumn{1}{c|}{31.91} & 29.93 \\
\multicolumn{1}{c|}{ChartGemma} & 50.48 & 38.21 & 32.10 & \multicolumn{1}{c|}{39.89} & 28.27 & 24.13 & 28.10 & 48.00 & 41.80 & 43.40 & \multicolumn{1}{c|}{\underline{42.47}} & 38.46 \\
\rowcolor{blue!5}\multicolumn{1}{c|}{ChartMoE (Ours)} & \textbf{71.44} & \textbf{51.57} & \underline{52.80} & \multicolumn{1}{c|}{\textbf{56.31}} & \textbf{38.40} & \underline{24.13} & \textbf{40.20} & \textbf{62.67} & \textbf{58.00} & \textbf{49.20} & \multicolumn{1}{c|}{\textbf{55.58}} & \textbf{51.67} \\ \bottomrule[1.5pt]
\end{tabular}
}
\vspace{-20pt}
\label{tab_chartbench}
\end{table*}

\subsection{Main Results}
\vspace{-5pt}
\textbf{ChartQA.}
Tab.\ref{tab_chartqa} presents detailed comparisons of \name\ on ChartQA. \name\ significantly improves the baseline (InternlmXC-v2) performance (72.00\% vs. 84.64\%, +12.64\%$\uparrow$ in Acc.@0.05). Compared to previous SOTA (TinyChart+PoT @768 pixel), \name\ consistently surpasses it across all metrics. The PoT effectively enhances the mathematical reasoning capabilities, which is a common shortfall in current MLLMs. \name\ integrates better with PoT, indicating that it accurately extracts fundamental elements from charts. \name\ shows more significant improvement on \textit{Human} part, especially after incorporating PoT, where the questions are more computationally complex and challenging. \revise{Notably, our error analysis in the \textit{Augmented} part reveals that many errors stem from limitations of the evaluation criteria, i.e., string matching. For instance, it is marked incorrect if the prediction is \textit{It is between 2003 and 2005} and the ground truth is \textit{(2003, 2005)}}. Forcing performance improvement may lead to model overfitting.

\textbf{ChartBench.}
Tab.~\ref{tab_chartbench} presents detailed comparisons of \name\ on ChartBench. None of the models, including our \name, undergo supervised fine-tuning on the ChartBench trainset to ensure fair experimental comparison. Chart-specific models typically underperform due to limited generalization, which fails to manage the annotated charts effectively ($<10$\%). Deplot shows a distinct advantage over these types of models (27.62\%) with the assistance of LLaVA-v1.6. The baseline (InternlmXC-v2) demonstrates strong generalization on ChartBench (48.41\%), which may benefit from pre-training instructions designed for unannotated charts. Without additional design, \name\ improves the baseline performance comprehensively (48.41\% vs. 51.67\%), especially on extra chart types (39.72\% vs. 55.58\%, +15.86\%$\uparrow$).

\textbf{ChartFC \& ChartCheck.}
Tab.~\ref{tab_chartfc} compares \name\ on the synthetic ChartFC and real-world ChartCheck. \name\ consistently outperforms SOTA (e.g., ChartGemma +4.4\%$\uparrow$ on ChartFC) and significantly improves the performance compared to InternlmXC-v2 (+6.83\%$\uparrow$ and +8.76\%$\uparrow$ on ChartCheck T1 and T2, respectively). Note that this is implemented without using training data for supervised fine-tuning, demonstrating \name's strong generalization capabilities.

\begin{minipage}{0.34\linewidth}
    \centering
    \resizebox{\linewidth}{!}{
    \setlength{\tabcolsep}{5pt}
    \begin{tabular}{@{}c|c|cc@{}}
    \toprule[1.5pt]
    \multirow{2}{*}{Models} & \multirow{2}{*}{ChartFC} & \multicolumn{2}{c}{ChartCheck} \\ \cmidrule(l){3-4} 
     &  & T1 & T2 \\ \midrule
    PaliGemma & 58.26 & 67.34 & 68.50 \\
    LLaVA-v1.5$\dagger$ & 61.28 & 70.22 & 70.03 \\
    InternlmXC-v2 & 65.93 & 72.04 & 70.44 \\
    ChartInstruct-LLama2 & 69.57 & 70.11 & 68.80 \\
    ChartInstruct-FlanT5XL & 70.27 & 72.03 & 73.80 \\
    ChartGemma & 70.33 & 71.50 & 74.31 \\
    \rowcolor{blue!5} ChartMoE (Ours) & \textbf{74.73} & \textbf{78.87} & \textbf{79.20} \\ \bottomrule[1.5pt]
    \end{tabular}
    }
    \captionsetup{skip=5pt}
    \captionof{table}{The accuracy performance on \textbf{\textit{ChartFC}} and \textbf{\textit{ChartCheck}}. $\dagger$: tuning with ChartGemma instructions.}
    \label{tab_chartfc}
\end{minipage}
\hfill
\begin{minipage}{0.28\linewidth}
    \centering
    \includegraphics[width=\linewidth]{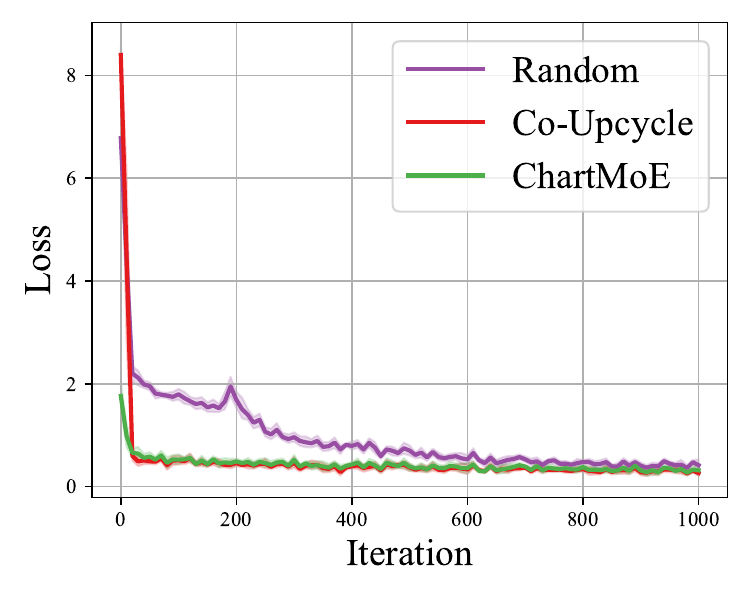}
    \captionsetup{skip=0pt}
    \captionof{figure}{Training loss of different initialization.}
    \label{fig_loss}
\end{minipage}
\hfill
\begin{minipage}{0.34\linewidth}
    \centering
    \includegraphics[width=\linewidth]{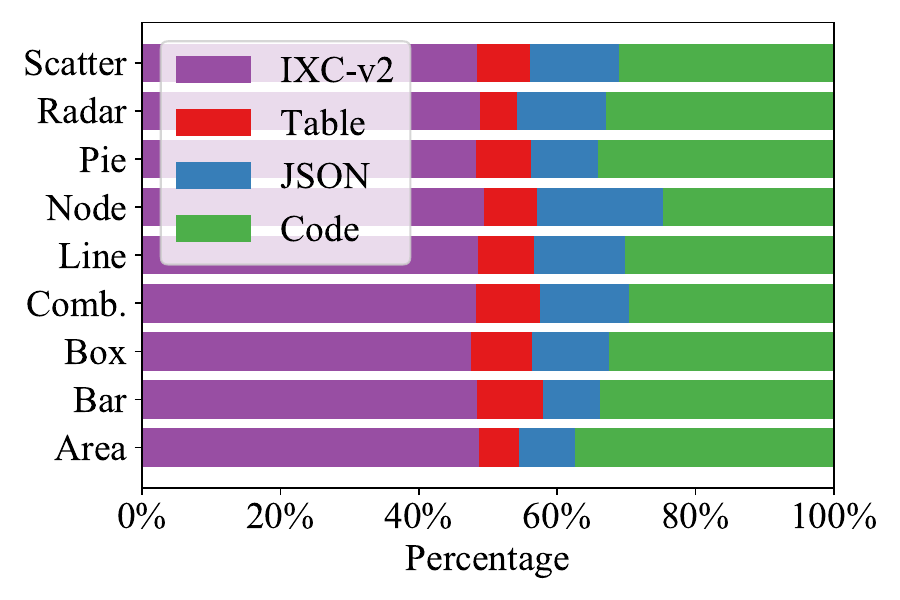}
    \captionsetup{skip=0pt}
    \captionof{figure}{Top-2 selected expert distribution on ChartBench.}
    \label{fig_topk_expert}
\end{minipage}

\section{Further Study}

\subsection{Model Architecture Ablation}
\vspace{-5pt}
We investigate the impact of three factors on our MoE connector: the number of experts, the number of activated experts, and the expert initialization manner. All the experiments are conducted with ChartQA training data and evaluated on ChartQA test split with relax accuracy metric.

\textbf{Effect of the Expert Initialization Manner.} 
The initialization strategy plays a crucial role in determining the performance of the MoE connector. Effective initialization is essential to ensure that each expert performs its designated function optimally. As illustrated in Tab.~\ref{tab_model_ablation} row 1-3, we explore the impact of 3 initialization strategies for the MoE connector. Random initialization serves as a baseline but struggles with convergence (refer to Fig.~\ref{fig_loss}), resulting in a suboptimal accuracy of 73.20\% at Acc.@0.05. Following CuMo~\cite{cumo}, we employ the Co-Upcycle strategy by replicating the \textit{table-JSON-code} aligned connector for all experts. Given the same starting point, this approach lacks expert diversity, which limits its effectiveness, resulting in an accuracy of 77.48\% at Acc.@0.05. In contrast, our initialization assigns distinct parameters to each expert. This tailored approach enables each expert to capitalize on its specific strengths, resulting in the highest performance, achieving 78.76\% in Acc.@0.05.

\textbf{Effect of Number of Experts and Activated Experts.} 
As shown in rows 3-4 of Tab.~\ref{tab_model_ablation}, we compare \name\ configurations with 4 and 8 experts, keeping 2 experts activated. The 8 experts are initialized in pairs using the 4 methods illustrated in Fig.~\ref{fig_train_pipeline}c. \name\ achieves 78.76\% in Acc.@0.05 with 4 experts, which is slightly higher than the 78.60\% achieved with 8 experts, showing a marginal increase of +0.16\%. In rows 4-5, we compare the performance of configurations with 2 and 4 activated experts, finding similar results: 78.60\% vs. 78.64\% in Acc.@0.05. This analysis suggests that merely increasing the number of experts or the activation of experts does not guarantee improved performance. The configuration with 4 experts and 2 activated experts effectively balances complexity and performance, making it a suitable choice for \name.

\subsection{Training Strategy Ablation}
\vspace{-5pt}
We analyze the impact of the training strategy across alignment and supervised fine-tuning stages. We use InternlmXC-v2 with ChartQA fine-tuning as our baseline, maintaining the same hyperparameters as the chart-specific annealing tuning stage.

\begin{table}
    \begin{minipage}{0.52\linewidth}
        \centering
        \captionsetup{skip=0pt}
        \caption{Ablation study on \name\ architecture w.r.t. the total / activated / initialization of connector experts. All experiments are conducted on ChartQA.}
        \resizebox{\linewidth}{!}{
        \setlength{\tabcolsep}{10pt}
        \begin{tabular}{@{}cccccc@{}}
        \toprule[1.5pt]
        \multicolumn{1}{c|}{\multirow{2}{*}{\makecell{Total\\Experts}}} & \multicolumn{1}{c|}{\multirow{2}{*}{\makecell{Activated\\Experts}}} & \multicolumn{1}{c|}{\multirow{2}{*}{\makecell{Experts\\Initialization}}} & \multicolumn{3}{c}{Relax Acc @0.05} \\ \cmidrule(l){4-6} 
        \multicolumn{1}{c|}{} & \multicolumn{1}{c|}{} & \multicolumn{1}{c|}{} & Human & Aug. & Avg. \\ \midrule
        \multicolumn{1}{c|}{4} & \multicolumn{1}{c|}{2} & \multicolumn{1}{c|}{Random Init.} & 59.68 & 86.72 & 73.20 \\ \midrule
        \multicolumn{1}{c|}{4} & \multicolumn{1}{c|}{2} & \multicolumn{1}{c|}{Random Align} & 62.32
         & 88.88 & 75.60 \\ \midrule
        \multicolumn{1}{c|}{4} & \multicolumn{1}{c|}{2} & \multicolumn{1}{c|}{Co-Upcycle Init.} & 64.96 & 90.00 & 77.48 \\ \midrule
        \rowcolor{blue!5}\multicolumn{1}{c|}{4} & \multicolumn{1}{c|}{2} & \multicolumn{1}{c|}{Diversely Align} & \textbf{67.92} & 89.60 & \textbf{78.76} \\ \midrule
        \multicolumn{1}{c|}{8} & \multicolumn{1}{c|}{2} & \multicolumn{1}{c|}{Diversely Align} & 67.20 & \textbf{90.00} & 78.60 \\ \midrule
        \multicolumn{1}{c|}{8} & \multicolumn{1}{c|}{4} & \multicolumn{1}{c|}{Diversely Align} & 67.68 & 89.60 & 78.64\\ \bottomrule[1.5pt]
        \end{tabular}
        }
        \label{tab_model_ablation}
    \end{minipage}
    \hfill
    \begin{minipage}{0.46\linewidth}
        \centering
        \captionsetup{skip=0pt}
        \caption{Ablation study on the proposed training strategy and connector architecture on the alignment, high-quality knowledge learning, and chart-specific anneal tuning stages.}
        \resizebox{\linewidth}{!}{
        \setlength{\tabcolsep}{5pt}
        \begin{tabular}{@{}cccc@{}}
        \toprule[1.5pt]
        \multicolumn{1}{c|}{\multirow{2}{*}{\name\ Recipe}} & \multicolumn{3}{c}{Relax Acc @0.05}  \\ \cmidrule(l){2-4} 
        \multicolumn{1}{c|}{} & Human & Aug. & \multicolumn{1}{c}{Avg.} \\ \midrule
        \multicolumn{1}{l|}{Baseline: InternlmXC-v2 + ChartQA} & 63.68 & 87.68 & \multicolumn{1}{c}{75.68} \\ \midrule
        \multicolumn{1}{l|}{+ \textit{table-JSON-code} Aligned Connector} & 64.24 & 90.16 & \multicolumn{1}{c}{77.20} \\ \midrule
        \multicolumn{1}{l|}{+ \textit{Top2-in-4} ChartMoE Connector} & 67.92 & 89.60 & \multicolumn{1}{c}{78.76} \\ \midrule
        \multicolumn{1}{l|}{+ \textit{MMC} High-Quality Knowledge Learning} & 67.84 & 90.24 & \multicolumn{1}{c}{79.04} \\ \midrule
        \rowcolor{blue!5}\multicolumn{1}{l|}{+ \textit{ChartGemma} Instructions} & \textbf{71.36} & \textbf{91.04} & \multicolumn{1}{c}{\textbf{81.20}}\\ 
        \bottomrule[1.5pt]
        \end{tabular}
        }
        \label{tab_data_ablation}
    \end{minipage}
\vspace{-5pt}
\end{table}

\begin{table}
    \begin{minipage}{0.39\linewidth}
        \centering
        \captionsetup{skip=0pt}
        \caption{Ablation study on the expert of MoE connector. We ignore the gating network and adopt specific expert output.}
        \resizebox{\linewidth}{!}{
        \setlength{\tabcolsep}{12pt}
        \begin{tabular}{@{}c|ccc@{}}
        \toprule[1.5pt]
        \multirow{2}{*}{Connector} & \multicolumn{3}{c}{Relax Acc @0.05} \\ \cmidrule(l){2-4} 
         & Human & Aug. & Avg. \\ \midrule
        Expert 0 (Vanilla) & \textbf{69.76} & \textbf{89.84} & \textbf{79.80} \\
        Expert 1 (Table) & 63.60 & 89.12 & 76.36 \\
        Expert 2 (JSON) & 60.64 & 82.48 & 71.56 \\
        Expert 3 (Code) & \underline{66.88} & \underline{89.36} & \underline{78.12} \\ \bottomrule[1.5pt]
        \end{tabular}
        }
        \label{tab_expert_ablation}
    \end{minipage}
    \hfill
    \begin{minipage}{0.59\linewidth}
        \centering
        \captionsetup{skip=0pt}
        \caption{Ablation study on alignment pre-training tasks. We adopt different alignment tasks for baseline (linear connector) and further conduct supervised fine-tuning on the ChartQA train set.}
        \resizebox{\linewidth}{!}{
        \setlength{\tabcolsep}{12pt}
        \begin{tabular}{@{}c|ccc|ccc@{}}
        \toprule[1.5pt]
        \multirow{2}{*}{Alignment} & \multicolumn{3}{c|}{\textit{w/o} ChartQA SFT} & \multicolumn{3}{c}{\textit{w/i} ChartQA SFT} \\ \cmidrule(l){2-7} 
         & Human & Aug. & Avg. & Human & Aug. & Avg. \\ \midrule
        Vanilla & \textbf{62.72} & \textbf{81.28} & \textbf{72.00} & 63.68 & 87.68 & 75.68 \\
        Table & \underline{51.28} & \underline{71.76} & \underline{61.52} & 63.92 & 89.28 & 76.60 \\
        JSON & 44.40 & 65.12 & 54.76 & \textbf{64.88} & \underline{89.84} & \textbf{77.36} \\
        Code & 50.16 & 71.20 & 60.68 & \underline{64.24} & \textbf{90.16} & \underline{77.20} \\ \bottomrule[1.5pt]
        \end{tabular}
        }
        \label{tab_align_ablation}
    \end{minipage}
\vspace{-5pt}
\end{table}

\textbf{Effect of Alignment Strategy.}
As shown in rows 1-3 of Tab. \ref{tab_data_ablation}, translating the chart image into structural text formats such as table, JSON, and code during the alignment stage significantly enhances performance in downstream chart understanding tasks. After applying \textit{table-JSON-code} alignment, the model achieves 77.20\% in Acc.@0.05, representing a notable improvement (+1.52\%$\uparrow$). When combined with our proposed MoE connector, performance further increases to 78.76\%, a total gain of +3.08\%$\uparrow$ in Acc.@0.05.

\textbf{Effect of Supervise Fine-Tuning Strategy.}
As shown in rows 4-5 of Tab.~\ref{tab_data_ablation}, we divide the supervised fine-tuning stage into two phases. By incorporating high-quality knowledge learning using the MMC dataset, \name\ achieves 79.04\% Acc.@0.05, reflecting a 3.36\% improvement. In the chart-specific annealing tuning phase, we introduce ChartGemma data to enhance the model's reasoning and PoT capabilities, leading the model to peak performance (81.20\%, +5.52\%$\uparrow$).

\subsection{In-depth Analysis}
\vspace{-5pt}

\begin{figure*}[t!]
    % \hsize=\textwidth
    \centering
      \begin{overpic}[width=1\linewidth, grid=False]{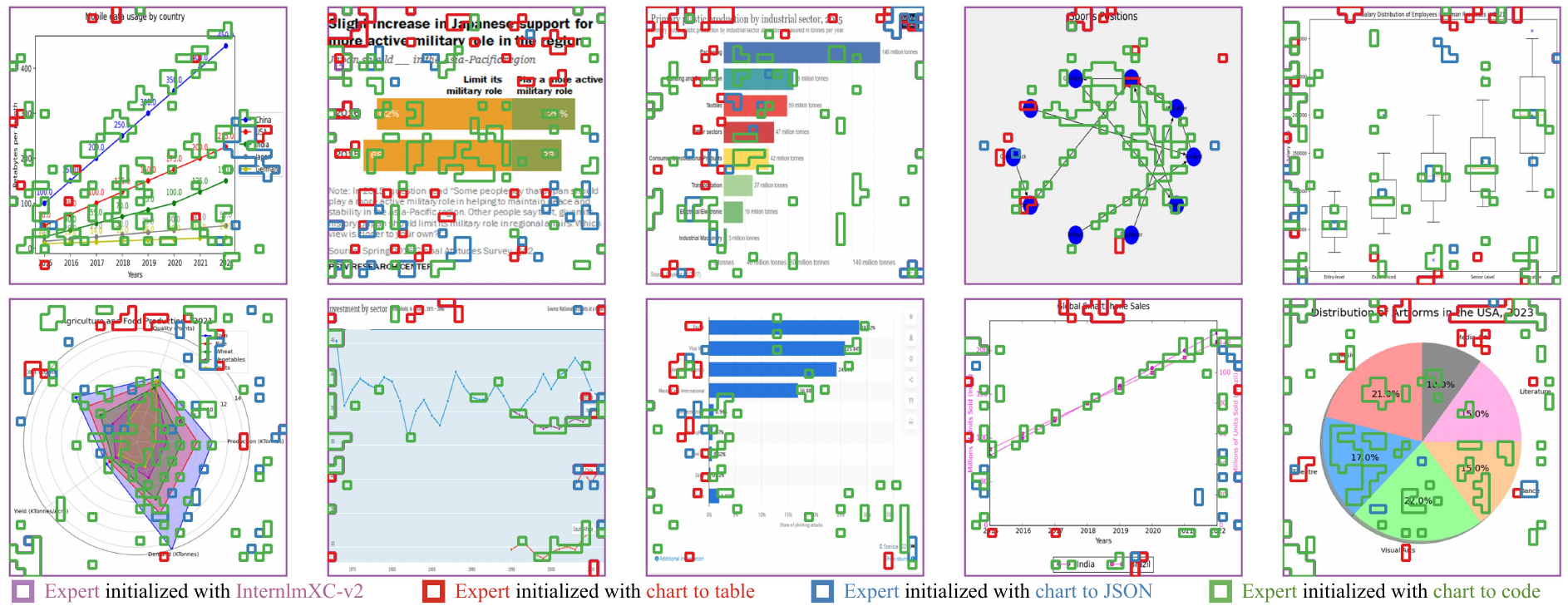}
      \end{overpic}
      \captionsetup{skip=0pt}
      \caption{Visualizations of top-1 expert selection. Only the boundaries of the merged tokens are plotted.}
      \vspace{-15pt}
    \label{fig_token_wise_expert}
\end{figure*}

\textbf{Effect of the Each Expert.} 
To explore the role of each expert in \name, we bypass the gating network and manually select the output of specific experts. As shown in Tab.~\ref{tab_expert_ablation}, E0 performs the best (79.80\%), which is consistent with the distribution in Fig.~\ref{fig_topk_expert}. However, this doesn't mean other experts lack relevance, which may offer crucial insights at key moments (Fig.~\ref{fig_token_wise_expert}).

\textbf{Effect of Alignment Task.}
As shown in Tab.~\ref{tab_align_ablation}, we explore various alignment tasks based on the linear connector. After alignment, the performance on ChartQA declines compared to the baseline. However, the aligned model exhibits a substantial improvement after supervised fine-tuning on the ChartQA train split, which is consistent with previous observations~\cite{ChartAst, ChartReformer}. Specifically, the JSON and code tasks exhibit remarkable improvement over the table.

\begin{figure}[t]
    \centering
    \includegraphics[width=\textwidth]{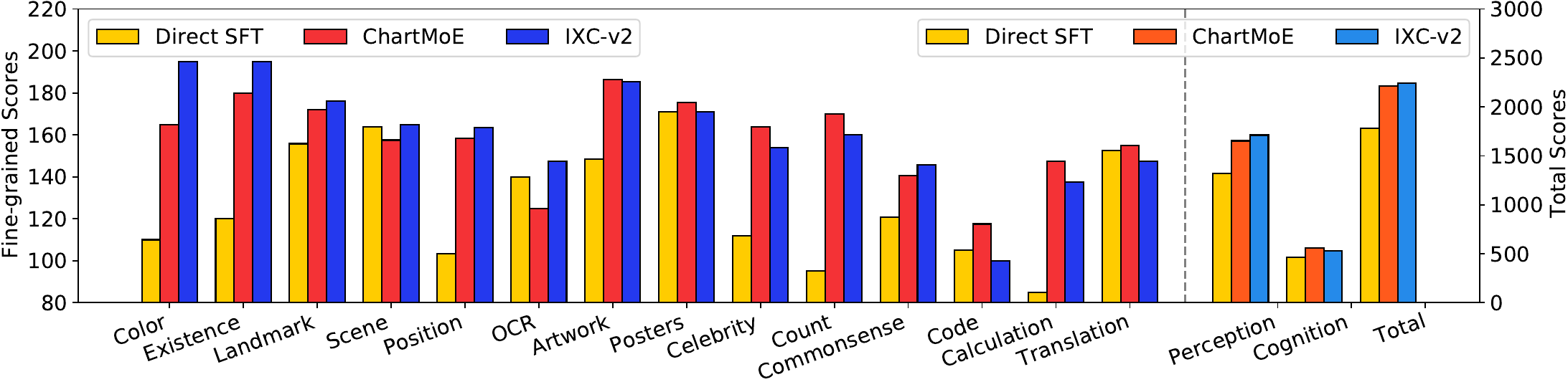}
    \captionsetup{skip=2pt}
    \caption{The performance on the general VQA tasks (MME~\cite{MME}). With supervised fine-tuning on extensive chart-structured data, the directly tuned IXC-v2 shows a significant performance drop, while \name\ maintains a satisfying performance by keeping the vanilla connector as the expert in MoE.}
    \vspace{-10pt}
    \label{tab_mme}
\end{figure}

\begin{figure}[t]
\centering
\begin{minipage}[t]{0.49\textwidth}
    \centering
    \includegraphics[width=\linewidth]{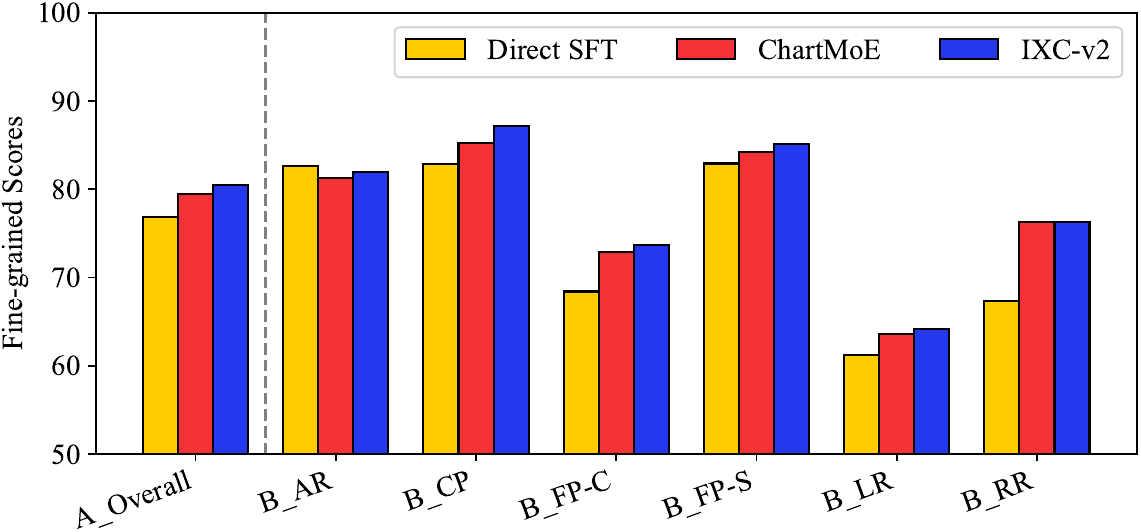}
    \captionsetup{skip=0pt}
    \captionof{figure}{The performance on the general VQA tasks (MMBench~\cite{MMBench}). Please refer to its paper for each task's details. The observations and conclusions are consistent with the MME benchmark.}
    \label{fig_mmb}
\end{minipage}
\hfill
\begin{minipage}[t]{0.47\textwidth} % [t] 用于垂直对齐顶部
    \centering
    \includegraphics[width=\linewidth]{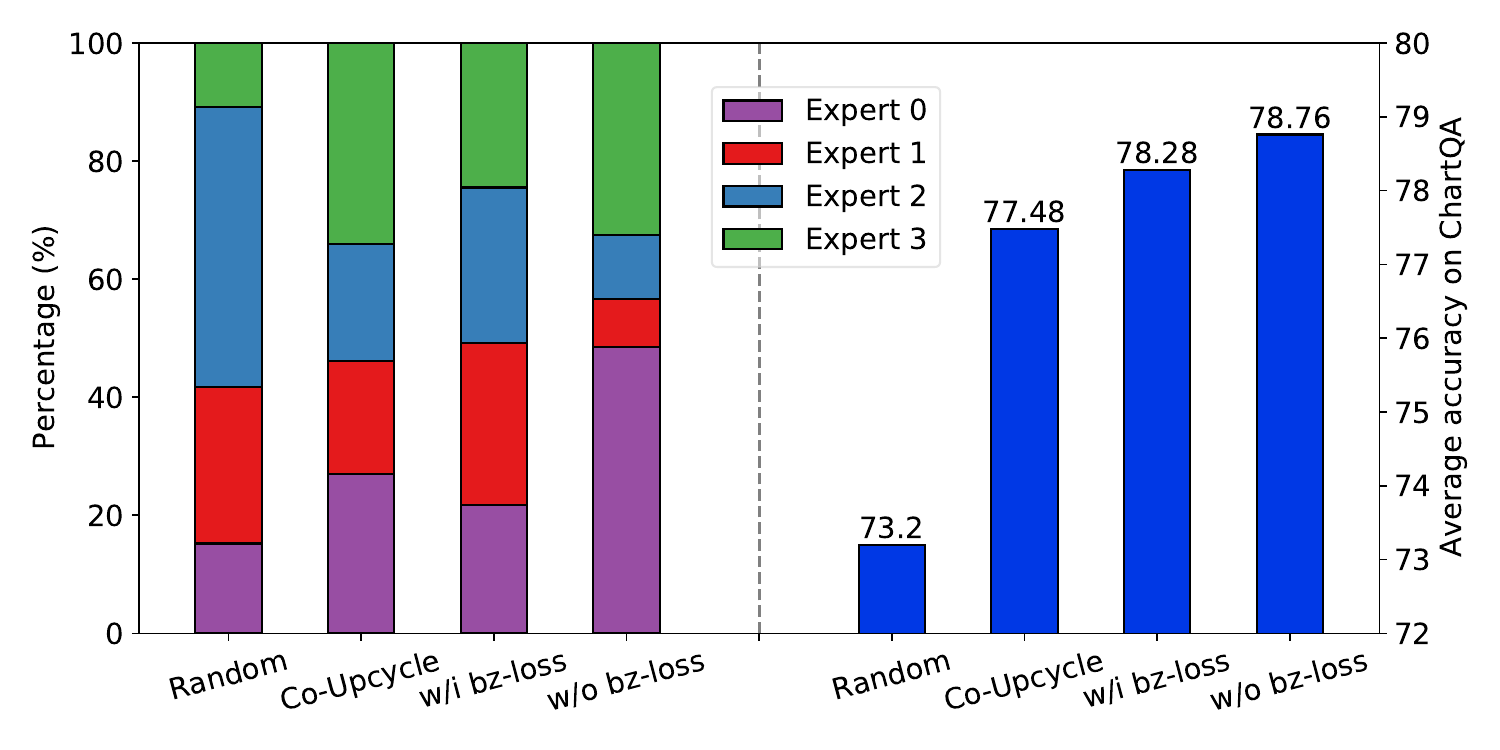}
    \captionsetup{skip=0pt}
    \captionof{figure}{The performance with/without bz-loss on ChartQA. Left: The bz-loss leads to more even expert selections. Right: A more balanced distribution does not yield better performance.}
    \label{fig_balloss}
\end{minipage}
\vspace{-10pt}
\end{figure}

\textbf{Expert Distribution Visualization.}
As shown in Fig.~\ref{fig_topk_expert}\&~\ref{fig_token_wise_expert}, we visualize the expert distribution in the MoE connector on the ChartBench test set. We designate the vanilla connector as E0, while E1-3 corresponds to connectors aligned with tables, JSON, and code. As depicted in Fig.~\ref{fig_topk_expert}, the trend is consistent across different chart types, with E0 and E3 being the most frequently selected connectors. The expert selection shows no extreme bias, as even the least chosen, E1, accounts for over 10\%. We further visualize the expert selection for each image token, revealing the preferences of each expert. As shown in Fig.~\ref{fig_token_wise_expert}, E0 is the primary choice for background tokens, explaining its dominance in Fig.~\ref{fig_topk_expert}. E1 and E2 are more frequently chosen by tokens from titles, axis labels, or legends, as these elements are commonly found in tables and JSON files. \name\ tends to use E3 to focus on the data points and visual elements within the chart, e.g., data points on the line, digital text, and edges in a node chart. These components are essential for accurately re-drawing the charts.

\textbf{Performance on General Tasks.}
While ChartMoE is designed to enhance chart understanding, it does not compromise other capabilities, e.g., instruction following and object recognition. In Fig.~\ref{tab_mme}\&~\ref{fig_mmb}, we show the comparisons of directly fine-tuned InternlmXC-v2 (short for Direct SFT) with data from Tab.~\ref{tab_dataset} and the baseline (short for IXC-v2) on general benchmarks~\cite{MME, MMBench}. The direct SFT model shows diminished general capabilities. In contrast, ChartMoE preserves it nearly intact by retaining the original connector as one of its experts.

\textbf{Effect of Balance Loss in MoE.}
The standard MoE~\cite{stmoe} employs balanced loss and router z-loss (short for bz-loss) to prevent certain experts from dominating the model training. In Fig.~\ref{fig_balloss}, we compare the effects of with and without bz-loss. While bz-loss promotes a more equitable selection of experts, it fails to enhance ChartMoE's performance further. As shown in Fig.~\ref{fig_token_wise_expert}, the expert initialization in ChartMoE results in each expert having its own preference for visual token selection (refer to Appendix \ref{apdx_sec_visual_token} for detail). Consequently, the bz-loss might hinder the model's convergence to the optimal point because the distribution of visual tokens is inherently imbalanced.

\section{Conclusion}
We introduce \name, a multi-task aligned and instruction-tuned MLLM designed for complex chart understanding and reasoning. 
%$
We replace the linear connector with the MoE architecture and initialize each expert with parameters derived from different alignment tasks.
We further present the \alignname \ dataset, a synthetic collection of nearly 1 million table-json-code-chart quadruples, to facilitate alignment training across different experts. 
This approach preserves the strengths of each alignment task, ensuring efficient training and superior model performance. 
ChartMoE outperforms the previous state-of-the-art on several benchmarks by a large margin and excels in real-world applications such as chart question answering, translation, and editing.
Please refer to Appendix \ref{apdx_repro} for the reproducibility statement.

\section{ACKNOWLEDGMENTS}
This work was supported by the National Key R\&D Program of China (2022YFB4701400/470140\\2), SSTIC Grant(KJZD20230923115106012, KJZD20230923114916032, GJHZ202402181136040\\08), and Beijing Key Lab of Networked Multimedia.

\bibliography{iclr2025_conference}

\begin{thebibliography}{69}
\providecommand{\natexlab}[1]{#1}
\providecommand{\url}[1]{\texttt{#1}}
\expandafter\ifx\csname urlstyle\endcsname\relax
  \providecommand{\doi}[1]{doi: #1}\else
  \providecommand{\doi}{doi: \begingroup \urlstyle{rm}\Url}\fi

\bibitem[Akhtar et~al.(2023{\natexlab{a}})Akhtar, Cocarascu, and Simperl]{chartfc}
Mubashara Akhtar, Oana Cocarascu, and Elena Paslaru~Bontas Simperl.
\newblock Reading and reasoning over chart images for evidence-based automated fact-checking.
\newblock \emph{arXiv preprint:2301.11843}, 2023{\natexlab{a}}.

\bibitem[Akhtar et~al.(2023{\natexlab{b}})Akhtar, Subedi, Gupta, Tahmasebi, Cocarascu, and Simperl]{chartcheck}
Mubashara Akhtar, Nikesh Subedi, Vivek Gupta, Sahar Tahmasebi, Oana Cocarascu, and Elena Simperl.
\newblock Chartcheck: An evidence-based fact-checking dataset over real-world chart images.
\newblock \emph{arXiv preprint:2311.07453}, 2023{\natexlab{b}}.

\bibitem[Alayrac et~al.(2022)Alayrac, Donahue, Luc, et~al.]{Flamingo}
Jean-Baptiste Alayrac, Jeff Donahue, Pauline Luc, et~al.
\newblock Flamingo: a visual language model for few-shot learning.
\newblock In \emph{proceedings of NeurIPS}, volume~35, pp.\  23716--23736, 2022.

\bibitem[Alibaba(2024)]{Qwen25}
Alibaba.
\newblock Qwen-vl 2.5: A multimodal large language model.
\newblock \url{https://tongyi.aliyun.com/qianwen}, 2024.
\newblock Accessed: 2024-09-17.

\bibitem[Bai et~al.(2023{\natexlab{a}})Bai, Bai, Chu, et~al.]{qwen}
Jinze Bai, Shuai Bai, Yunfei Chu, et~al.
\newblock Qwen technical report.
\newblock \emph{arXiv preprint:2309.16609}, 2023{\natexlab{a}}.

\bibitem[Bai et~al.(2023{\natexlab{b}})Bai, Bai, Yang, et~al.]{qwen-vl}
Jinze Bai, Shuai Bai, Shusheng Yang, et~al.
\newblock Qwen-vl: A frontier large vision-language model with versatile abilities.
\newblock \emph{arXiv preprint:2308.12966}, 2023{\natexlab{b}}.

\bibitem[Beyer et~al.(2024)Beyer, Steiner, Pinto, et~al.]{paligemma}
Lucas Beyer, Andreas Steiner, Andr{\'e}~Susano Pinto, et~al.
\newblock Paligemma: A versatile 3b vlm for transfer.
\newblock \emph{arXiv preprint:2407.07726}, 2024.

\bibitem[Brown et~al.(2020)Brown, Mann, Ryder, et~al.]{GPT-3}
Tom Brown, Benjamin Mann, Nick Ryder, et~al.
\newblock Language models are few-shot learners.
\newblock In \emph{proceedings of NeurIPS}, volume~33, pp.\  1877--1901, 2020.

\bibitem[Carbune et~al.(2024)Carbune, Mansoor, Liu, et~al.]{ChartPaLI}
Victor Carbune, Hassan Mansoor, Fangyu Liu, et~al.
\newblock Chart-based reasoning: Transferring capabilities from llms to vlms.
\newblock In \emph{proceedings of NAACL}, 2024.

\bibitem[Chen et~al.(2024)Chen, Kong, Wei, Liu, Ge, Zhao, Sun, Han, and Zhang]{OneChart}
Jinyue Chen, Lingyu Kong, Haoran Wei, Chenglong Liu, Zheng Ge, Liang Zhao, Jianjian Sun, Chunrui Han, and Xiangyu Zhang.
\newblock Onechart: Purify the chart structural extraction via one auxiliary token.
\newblock \emph{arXiv preprint:2404.09987}, 2024.

\bibitem[Chen et~al.(2023)Chen, Zhu, Shen, et~al.]{Minigptv2}
Jun Chen, Deyao Zhu, Xiaoqian Shen, et~al.
\newblock Minigpt-v2: large language model as a unified interface for vision-language multi-task learning.
\newblock \emph{arXiv preprint:2310.09478}, 2023.

\bibitem[Dao(2024)]{flashattention2}
Tri Dao.
\newblock Flash{A}ttention-2: Faster attention with better parallelism and work partitioning.
\newblock In \emph{International Conference on Learning Representations (ICLR)}, 2024.

\bibitem[Dong et~al.(2024)Dong, Zhang, Zang, et~al.]{internlm-xcomposerv2}
Xiaoyi Dong, Pan Zhang, Yuhang Zang, et~al.
\newblock Internlm-xcomposer2: Mastering free-form text-image composition and comprehension in vision-language large model.
\newblock \emph{arXiv preprint:2401.16420}, 2024.

\bibitem[Fu et~al.(2023)Fu, Chen, Shen, et~al.]{MME}
Chaoyou Fu, Peixian Chen, Yunhang Shen, et~al.
\newblock Mme: A comprehensive evaluation benchmark for multimodal large language models.
\newblock \emph{arXiv preprint:2306.13394}, 2023.

\bibitem[Han et~al.(2023)Han, Zhang, Chen, et~al.]{Chartllama}
Yucheng Han, Chi Zhang, Xin Chen, et~al.
\newblock Chartllama: A multimodal llm for chart understanding and generation.
\newblock \emph{arXiv preprint:2311.16483}, 2023.

\bibitem[He et~al.(2016)He, Zhang, Ren, and Sun]{resnet}
Kaiming He, Xiangyu Zhang, Shaoqing Ren, and Jian Sun.
\newblock Deep residual learning for image recognition.
\newblock In \emph{proceedings of CVPR}, pp.\  770--778, 2016.

\bibitem[Hoque et~al.(2017)Hoque, Setlur, Tory, and Dykeman]{introtext}
Enamul Hoque, Vidya Setlur, Melanie Tory, and Isaac Dykeman.
\newblock Applying pragmatics principles for interaction with visual analytics.
\newblock \emph{IEEE TVCG}, 24\penalty0 (1):\penalty0 309--318, 2017.

\bibitem[Hsu et~al.(2021)Hsu, Giles, and Huang]{SciCap}
Ting-Yao Hsu, C~Lee Giles, and Ting-Hao'Kenneth' Huang.
\newblock Scicap: Generating captions for scientific figures.
\newblock In \emph{Findings of ACL}, 2021.

\bibitem[Hu et~al.(2024)Hu, Xu, Ye, Yan, Zhang, Zhang, Li, Zhang, Jin, Huang, et~al.]{DocOwl}
Anwen Hu, Haiyang Xu, Jiabo Ye, Ming Yan, Liang Zhang, Bo~Zhang, Chen Li, Ji~Zhang, Qin Jin, Fei Huang, et~al.
\newblock mplug-docowl 1.5: Unified structure learning for ocr-free document understanding.
\newblock \emph{arXiv preprint:2403.12895}, 2024.

\bibitem[Hu et~al.(2022)Hu, Shen, Wallis, et~al.]{lora}
Edward~J Hu, Yelong Shen, Phillip Wallis, et~al.
\newblock Lo{RA}: Low-rank adaptation of large language models.
\newblock In \emph{proceedings of ICLR}, 2022.

\bibitem[Huang et~al.(2024)Huang, Chan, Fung, Qiu, Zhou, Joty, Chang, and Ji]{awesome-chart}
Kung-Hsiang Huang, Hou~Pong Chan, Yi~R. Fung, Haoyi Qiu, Mingyang Zhou, Shafiq Joty, Shih-Fu Chang, and Heng Ji.
\newblock From pixels to insights: A survey on automatic chart understanding in the era of large foundation models, 2024.

\bibitem[Komatsuzaki et~al.(2023)Komatsuzaki, Puigcerver, Lee-Thorp, et~al.]{CoUpcycling}
Aran Komatsuzaki, Joan Puigcerver, James Lee-Thorp, et~al.
\newblock Sparse upcycling: Training mixture-of-experts from dense checkpoints.
\newblock In \emph{proceedings of ICLR}, 2023.

\bibitem[Lee et~al.(2023)Lee, Joshi, Turc, et~al.]{Pix2Str}
Kenton Lee, Mandar Joshi, Iulia~Raluca Turc, et~al.
\newblock Pix2struct: Screenshot parsing as pretraining for visual language understanding.
\newblock In \emph{proceedings of ICML}, pp.\  18893--18912, 2023.

\bibitem[Li et~al.(2024)Li, Wang, Zhu, Kuo, Xu, Chen, Jain, Shi, and Wen]{cumo}
Jiachen Li, Xinyao Wang, Sijie Zhu, Chia-Wen Kuo, Lu~Xu, Fan Chen, Jitesh Jain, Humphrey Shi, and Longyin Wen.
\newblock Cumo: Scaling multimodal llm with co-upcycled mixture-of-experts.
\newblock \emph{arXiv preprint:2405.05949}, 2024.

\bibitem[Li et~al.(2023)Li, Li, Savarese, et~al.]{BLIP2}
Junnan Li, Dongxu Li, Silvio Savarese, et~al.
\newblock {BLIP-2:} bootstrapping language-image pre-training with frozen image encoders and large language models.
\newblock In \emph{proceedings of ICML}, volume 202, pp.\  19730--19742, 2023.

\bibitem[Lin et~al.(2024)Lin, Tang, Ye, Cui, Zhu, Jin, Zhang, Ning, and Yuan]{moe-llava}
Bin Lin, Zhenyu Tang, Yang Ye, Jiaxi Cui, Bin Zhu, Peng Jin, Junwu Zhang, Munan Ning, and Li~Yuan.
\newblock Moe-llava: Mixture of experts for large vision-language models.
\newblock \emph{arXiv preprint arXiv:2401.15947}, 2024.

\bibitem[Lin et~al.(2023)Lin, Liu, Zhang, et~al.]{sphinx}
Ziyi Lin, Chris Liu, Renrui Zhang, et~al.
\newblock Sphinx: The joint mixing of weights, tasks, and visual embeddings for multi-modal large language models.
\newblock \emph{arXiv preprint:2311.07575}, 2023.

\bibitem[Liu et~al.(2023{\natexlab{a}})Liu, Eisenschlos, Piccinno, Krichene, et~al.]{DePlot}
Fangyu Liu, Julian~Martin Eisenschlos, Francesco Piccinno, Syrine Krichene, et~al.
\newblock Deplot: One-shot visual language reasoning by plot-to-table translation.
\newblock In \emph{Findings of ACL}, pp.\  10381--10399, 2023{\natexlab{a}}.

\bibitem[Liu et~al.(2023{\natexlab{b}})Liu, Piccinno, Krichene, et~al.]{MatCha}
Fangyu Liu, Francesco Piccinno, Syrine Krichene, et~al.
\newblock Matcha: Enhancing visual language pretraining with math reasoning and chart derendering.
\newblock In \emph{proceedings of ACL}, pp.\  12756--12770, 2023{\natexlab{b}}.

\bibitem[Liu et~al.(2023{\natexlab{c}})Liu, Wang, Yao, Chen, Song, Cho, Yacoob, and Yu]{MMC}
Fuxiao Liu, Xiaoyang Wang, Wenlin Yao, Jianshu Chen, Kaiqiang Song, Sangwoo Cho, Yaser Yacoob, and Dong Yu.
\newblock {MMC:} advancing multimodal chart understanding with large-scale instruction tuning.
\newblock In \emph{proceedings of ACL}, 2023{\natexlab{c}}.

\bibitem[Liu et~al.(2023{\natexlab{d}})Liu, Li, Wu, and Lee]{llava}
Haotian Liu, Chunyuan Li, Qingyang Wu, and Yong~Jae Lee.
\newblock Visual instruction tuning.
\newblock In \emph{proceedings of NeurIPS}, 2023{\natexlab{d}}.

\bibitem[Liu et~al.(2024{\natexlab{a}})Liu, Li, Li, Li, Zhang, Shen, and Lee]{llavanext}
Haotian Liu, Chunyuan Li, Yuheng Li, Bo~Li, Yuanhan Zhang, Sheng Shen, and Yong~Jae Lee.
\newblock Llava-next: Improved reasoning, ocr, and world knowledge, January 2024{\natexlab{a}}.
\newblock URL \url{https://llava-vl.github.io/blog/2024-01-30-llava-next/}.

\bibitem[Liu et~al.(2024{\natexlab{b}})Liu, Li, Li, et~al.]{Improvedllava}
Haotian Liu, Chunyuan Li, Yuheng Li, et~al.
\newblock Improved baselines with visual instruction tuning.
\newblock In \emph{proceedings of CVPR}, 2024{\natexlab{b}}.

\bibitem[Liu et~al.(2024{\natexlab{c}})Liu, Chen, Li, Fang, and Shen]{ChartThinker}
Mengsha Liu, Daoyuan Chen, Yaliang Li, Guian Fang, and Ying Shen.
\newblock Chartthinker: A contextual chain-of-thought approach to optimized chart summarization.
\newblock \emph{arXiv preprint:2403.11236}, 2024{\natexlab{c}}.

\bibitem[Liu et~al.(2023{\natexlab{e}})Liu, Duan, Zhang, et~al.]{MMBench}
Yuan Liu, Haodong Duan, Yuanhan Zhang, et~al.
\newblock Mmbench: Is your multi-modal model an all-around player?
\newblock \emph{arXiv preprint:2307.06281}, 2023{\natexlab{e}}.

\bibitem[Lu et~al.(2024)Lu, Liu, Zhang, Wang, Dong, Liu, Sun, Ren, Li, Yang, et~al.]{deepseek}
Haoyu Lu, Wen Liu, Bo~Zhang, Bingxuan Wang, Kai Dong, Bo~Liu, Jingxiang Sun, Tongzheng Ren, Zhuoshu Li, Hao Yang, et~al.
\newblock Deepseek-vl: towards real-world vision-language understanding.
\newblock \emph{arXiv preprint:2403.05525}, 2024.

\bibitem[Masry et~al.(2022)Masry, Long, Tan, et~al.]{ChartQA}
Ahmed Masry, Do~Xuan Long, Jia~Qing Tan, et~al.
\newblock Chartqa: A benchmark for question answering about charts with visual and logical reasoning.
\newblock In \emph{proceedings of ACL}, 2022.

\bibitem[Masry et~al.(2023)Masry, Kavehzadeh, Do, Hoque, and Joty]{unichart}
Ahmed Masry, Parsa Kavehzadeh, Xuan~Long Do, Enamul Hoque, and Shafiq Joty.
\newblock Unichart: A universal vision-language pretrained model for chart comprehension and reasoning.
\newblock In \emph{proceedings of EMNLP}, 2023.

\bibitem[Masry et~al.(2024{\natexlab{a}})Masry, Shahmohammadi, Parvez, Hoque, and Joty]{chartinstruct}
Ahmed Masry, Mehrad Shahmohammadi, Md~Rizwan Parvez, Enamul Hoque, and Shafiq Joty.
\newblock Chartinstruct: Instruction tuning for chart comprehension and reasoning.
\newblock \emph{arXiv preprint:2403.09028}, 2024{\natexlab{a}}.

\bibitem[Masry et~al.(2024{\natexlab{b}})Masry, Thakkar, Bajaj, Kartha, Hoque, and Joty]{chartgemma}
Ahmed Masry, Megh Thakkar, Aayush Bajaj, Aaryaman Kartha, Enamul Hoque, and Shafiq Joty.
\newblock Chartgemma: Visual instruction-tuning for chart reasoning in the wild.
\newblock \emph{arXiv preprint:2407.04172}, 2024{\natexlab{b}}.

\bibitem[Meng et~al.(2024)Meng, Shao, Lu, et~al.]{ChartAst}
Fanqing Meng, Wenqi Shao, Quanfeng Lu, et~al.
\newblock Chartassisstant: A universal chart multimodal language model via chart-to-table pre-training and multitask instruction tuning.
\newblock \emph{arXiv preprint:2401.02384}, 2024.

\bibitem[Methani et~al.(2020)Methani, Ganguly, Khapra, and Kumar]{PlotQA}
Nitesh Methani, Pritha Ganguly, Mitesh~M Khapra, and Pratyush Kumar.
\newblock Plotqa: Reasoning over scientific plots.
\newblock In \emph{proceedings of CVPR}, pp.\  1527--1536, 2020.

\bibitem[OpenAI(2023)]{GPT4}
OpenAI.
\newblock Gpt-4 technical report.
\newblock \emph{arXiv preprint:2303.08774}, 2023.

\bibitem[OpenAI(2024)]{gpt4o}
OpenAI.
\newblock Gpt-4o: A multimodal large language model.
\newblock \url{https://openai.com}, 2024.
\newblock Accessed: 2024-09-17.

\bibitem[Oquab et~al.(2023)Oquab, Darcet, Moutakanni, et~al.]{dinov2}
Maxime Oquab, Timoth{\'e}e Darcet, Th{\'e}o Moutakanni, et~al.
\newblock Dinov2: Learning robust visual features without supervision.
\newblock \emph{TMLR}, 2023.

\bibitem[Radford et~al.(2018)Radford, Narasimhan, Salimans, Sutskever, et~al.]{GPT-1}
Alec Radford, Karthik Narasimhan, Tim Salimans, Ilya Sutskever, et~al.
\newblock Improving language understanding by generative pre-training.
\newblock \emph{OpenAI blog}, 2018.

\bibitem[Radford et~al.(2021)Radford, Kim, Hallacy, et~al.]{clip}
Alec Radford, Jong~Wook Kim, Chris Hallacy, et~al.
\newblock Learning transferable visual models from natural language supervision.
\newblock In \emph{proceedings of ICML}, pp.\  8748--8763. PMLR, 2021.

\bibitem[Team et~al.(2024{\natexlab{a}})Team, Mesnard, Hardin, et~al.]{gemma}
Gemma Team, Thomas Mesnard, Cassidy Hardin, et~al.
\newblock Gemma: Open models based on gemini research and technology.
\newblock \emph{arXiv preprint:2403.08295}, 2024{\natexlab{a}}.

\bibitem[Team et~al.(2024{\natexlab{b}})Team, Mesnard, Hardin, et~al.]{llama3}
Gemma Team, Thomas Mesnard, Cassidy Hardin, et~al.
\newblock Gemma: Open models based on gemini research and technology.
\newblock \emph{arXiv preprint:2407.21783}, 2024{\natexlab{b}}.

\bibitem[Tong et~al.(2024{\natexlab{a}})Tong, Brown, Wu, Woo, Middepogu, Akula, Yang, Yang, Iyer, Pan, et~al.]{cambrian-1}
Shengbang Tong, Ellis Brown, Penghao Wu, Sanghyun Woo, Manoj Middepogu, Sai~Charitha Akula, Jihan Yang, Shusheng Yang, Adithya Iyer, Xichen Pan, et~al.
\newblock Cambrian-1: A fully open, vision-centric exploration of multimodal llms.
\newblock \emph{arXiv preprint arXiv:2406.16860}, 2024{\natexlab{a}}.

\bibitem[Tong et~al.(2024{\natexlab{b}})Tong, Liu, Zhai, Ma, LeCun, and Xie]{MMVP-MoF}
Shengbang Tong, Zhuang Liu, Yuexiang Zhai, Yi~Ma, Yann LeCun, and Saining Xie.
\newblock Eyes wide shut? exploring the visual shortcomings of multimodal llms.
\newblock In \emph{Proceedings of CVPR}, pp.\  9568--9578, 2024{\natexlab{b}}.

\bibitem[Touvron et~al.(2023)Touvron, Lavril, Izacard, et~al.]{llama}
Hugo Touvron, Thibaut Lavril, Gautier Izacard, et~al.
\newblock Llama: Open and efficient foundation language models.
\newblock \emph{arXiv preprint:2302.13971}, 2023.

\bibitem[Wang et~al.(2023)Wang, Golovneva, Aghajanyan, et~al.]{DOMINO}
Peifang Wang, Olga Golovneva, Armen Aghajanyan, et~al.
\newblock {DOMINO:} {A} dual-system for multi-step visual language reasoning.
\newblock \emph{arXiv preprint:2310.02804}, 2023.

\bibitem[Wei et~al.(2022)Wei, Wang, Schuurmans, et~al.]{CoT}
Jason Wei, Xuezhi Wang, Dale Schuurmans, et~al.
\newblock Chain-of-thought prompting elicits reasoning in large language models.
\newblock In \emph{proceedings of NeurIPS}, volume~35, pp.\  24824--24837, 2022.

\bibitem[Wu et~al.(2024)Wu, Yan, Luo, Wang, and Tang]{chartinsights}
Yifan Wu, Lutao Yan, Yuyu Luo, Yunhai Wang, and Nan Tang.
\newblock Evaluating task-based effectiveness of mllms on charts.
\newblock \emph{arXiv preprint:2405.07001}, 2024.

\bibitem[Xia et~al.(2024)Xia, Zhang, Ye, Yan, et~al.]{ChartVLM}
Renqiu Xia, Bo~Zhang, Hancheng Ye, Xiangchao Yan, et~al.
\newblock Chartx {\&} chartvlm: {A} versatile benchmark and foundation model for complicated chart reasoning.
\newblock \emph{arXiv preprint:2402.12185}, 2024.

\bibitem[Xu et~al.(2023)Xu, Du, Qi, Xu, Yuan, and Guo]{ChartBench}
Zhengzhuo Xu, Sinan Du, Yiyan Qi, Chengjin Xu, Chun Yuan, and Jian Guo.
\newblock Chartbench: A benchmark for complex visual reasoning in charts.
\newblock \emph{arXiv preprint:2312.15915}, 2023.

\bibitem[Xue et~al.(2024)Xue, Shu, Awadalla, Wang, Yan, Purushwalkam, Zhou, Prabhu, Dai, Ryoo, et~al.]{BLIP3}
Le~Xue, Manli Shu, Anas Awadalla, Jun Wang, An~Yan, Senthil Purushwalkam, Honglu Zhou, Viraj Prabhu, Yutong Dai, Michael~S Ryoo, et~al.
\newblock xgen-mm (blip-3): A family of open large multimodal models.
\newblock \emph{arXiv preprint:2408.08872}, 2024.

\bibitem[Yan et~al.(2024)Yan, Bhosale, Lal, Adhikari, and Doermann]{ChartReformer}
Pengyu Yan, Mahesh Bhosale, Jay Lal, Bikhyat Adhikari, and David~S. Doermann.
\newblock Chartreformer: Natural language-driven chart image editing.
\newblock \emph{arXiv preprint:2403.00209}, 2024.

\bibitem[Ye et~al.(2023{\natexlab{a}})Ye, Hu, Xu, et~al.]{Ureader}
Jiabo Ye, Anwen Hu, Haiyang Xu, et~al.
\newblock Ureader: Universal ocr-free visually-situated language understanding with multimodal large language model.
\newblock In \emph{Findings of ACL}, 2023{\natexlab{a}}.

\bibitem[Ye et~al.(2023{\natexlab{b}})Ye, Xu, Xu, et~al.]{Mplug}
Qinghao Ye, Haiyang Xu, Guohai Xu, et~al.
\newblock mplug-owl: Modularization empowers large language models with multimodality.
\newblock \emph{arXiv preprint:2304.14178}, 2023{\natexlab{b}}.

\bibitem[Ye et~al.(2023{\natexlab{c}})Ye, Xu, Ye, et~al.]{mPLUG2}
Qinghao Ye, Haiyang Xu, Jiabo Ye, et~al.
\newblock mplug-owl2: Revolutionizing multi-modal large language model with modality collaboration.
\newblock \emph{arXiv preprint:2311.04257}, 2023{\natexlab{c}}.

\bibitem[Zhai et~al.(2023)Zhai, Mustafa, Kolesnikov, and Beyer]{SigLIP}
Xiaohua Zhai, Basil Mustafa, Alexander Kolesnikov, and Lucas Beyer.
\newblock Sigmoid loss for language image pre-training.
\newblock In \emph{proceedings of ICCV}, pp.\  11975--11986, 2023.

\bibitem[Zhang et~al.(2024)Zhang, Hu, Xu, Yan, Xu, Jin, Zhang, and Huang]{TinyChart}
Liang Zhang, Anwen Hu, Haiyang Xu, Ming Yan, Yichen Xu, Qin Jin, Ji~Zhang, and Fei Huang.
\newblock Tinychart: Efficient chart understanding with visual token merging and program-of-thoughts learning.
\newblock \emph{arXiv preprint:2404.16635}, 2024.

\bibitem[Zhang et~al.(2023)Zhang, Wang, Cao, Xu, et~al.]{internlm-xcomposer}
Pan Zhang, Xiaoyi Dong~Bin Wang, Yuhang Cao, Chao Xu, et~al.
\newblock Internlm-xcomposer: A vision-language large model for advanced text-image comprehension and composition.
\newblock \emph{arXiv preprint:2309.15112}, 2023.

\bibitem[Zhang et~al.(2022)Zhang, Roller, Goyal, et~al.]{OPT}
Susan Zhang, Stephen Roller, Naman Goyal, et~al.
\newblock Opt: Open pre-trained transformer language models.
\newblock \emph{arXiv preprint:2205.01068}, 2022.

\bibitem[Zheng et~al.(2023)Zheng, Chiang, Sheng, et~al.]{vicuna}
Lianmin Zheng, Wei-Lin Chiang, Ying Sheng, et~al.
\newblock Judging llm-as-a-judge with mt-bench and chatbot arena.
\newblock In \emph{proceedings of NeurIPS}, 2023.

\bibitem[Zhuowan et~al.(2024)Zhuowan, Bhavan, Peng, and Shabnam]{LAMENDA}
Li~Zhuowan, Jasani Bhavan, Tang Peng, and Ghadar Shabnam.
\newblock Synthesize step-by-step: Tools, templates and llms as data generators for reasoning-based chart vqa.
\newblock In \emph{proceedings of CVPR}, 2024.

\bibitem[Zoph et~al.(2022)Zoph, Bello, Kumar, Du, Huang, Dean, Shazeer, and Fedus]{stmoe}
Barret Zoph, Irwan Bello, Sameer Kumar, Nan Du, Yanping Huang, Jeff Dean, Noam~M. Shazeer, and William Fedus.
\newblock St-moe: Designing stable and transferable sparse expert models.
\newblock \emph{arXiv preprint:2202.08906}, 2022.

\end{thebibliography}
\bibliographystyle{iclr2025_conference}

\small 

\clearpage
\appendix

\section{Additional experimental Settings and results}

\subsection{Top-2 experts distribution}
\begin{figure}[h]
    \begin{subfigure}[b]{0.25\textwidth}
        \includegraphics[width=\textwidth]{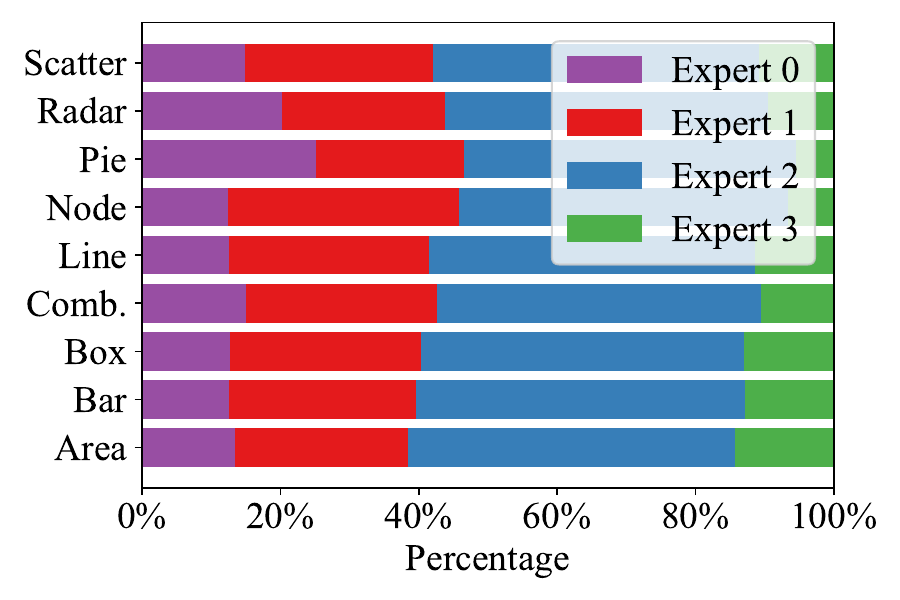}
        \captionsetup{skip=0pt}
        \caption{Random}
    \end{subfigure}%\
    \hfill
    \begin{subfigure}[b]{0.25\textwidth}
        \includegraphics[width=\textwidth]{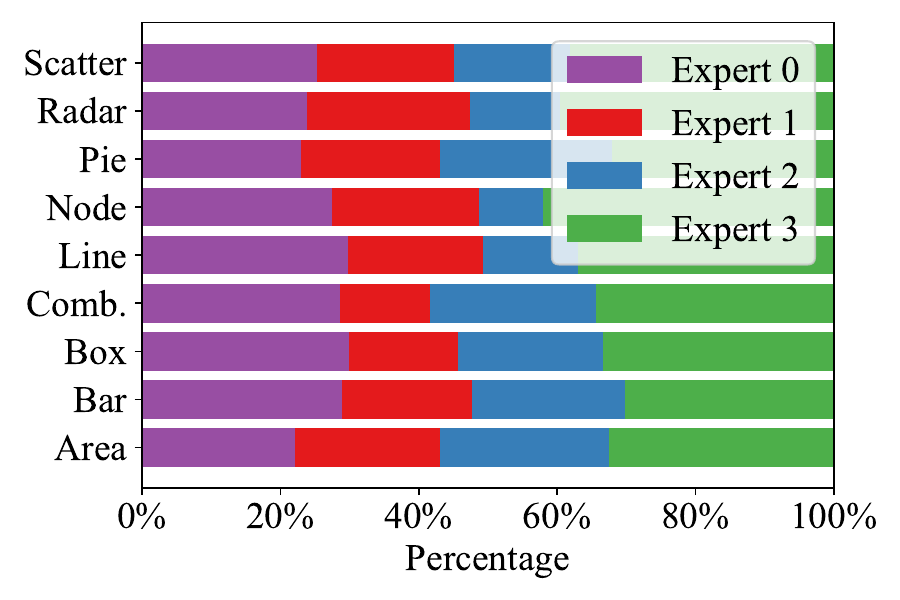}
        \captionsetup{skip=0pt}
        \caption{Co-Upcycle}
    \end{subfigure}%
    \begin{subfigure}[b]{0.25\textwidth}
        \includegraphics[width=\textwidth]{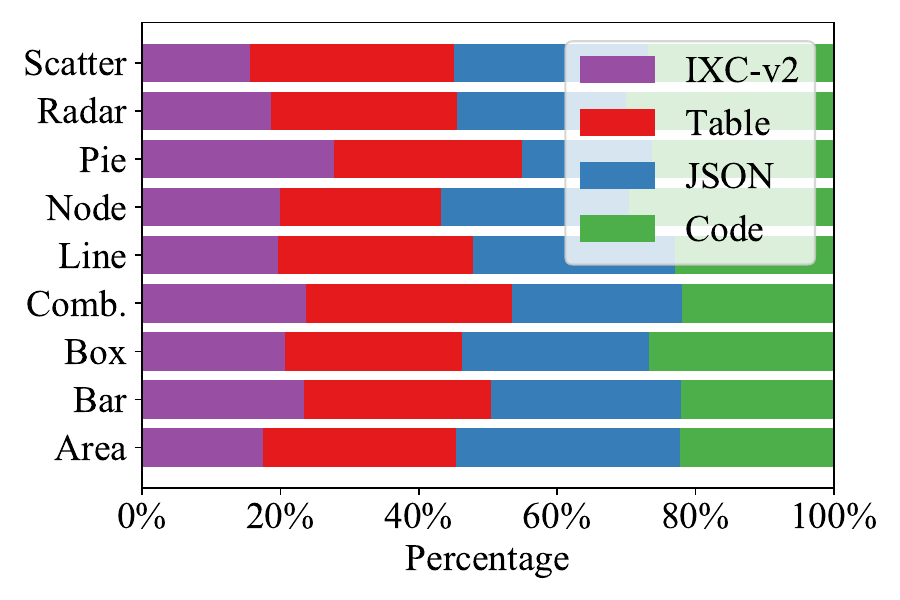}
        \captionsetup{skip=0pt}
        \caption{\name\ with bz-loss}
    \end{subfigure}%
    \hfill
    \begin{subfigure}[b]{0.25\textwidth}
        \includegraphics[width=\textwidth]{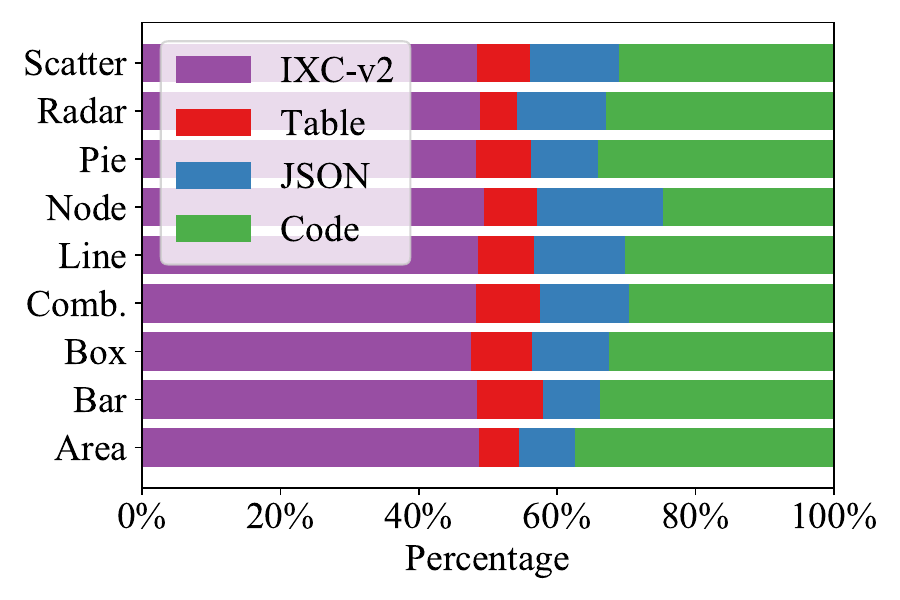}
        \captionsetup{skip=0pt}
        \caption{\name}
    \end{subfigure}%
    \centering
    % \captionsetup{skip=3pt}
    \caption{The distribution of Top-2 experts after supervised fine-tuning with three expert initialization methods. We calculate the proportion of the top 2 experts selected by the router on the ChartBench.}
    \label{fig_apdx_topk_expert}
\end{figure}

Our \name\ employs MoE connector expert parameters initialized with various alignment tasks. To investigate the impact of these initialization methods on model performance, we present the comparisons in Tab.~\ref{tab_data_ablation}\&~\ref{tab_expert_ablation}\&~\ref{tab_align_ablation} and Fig.~\ref{fig_loss}\&\ref{fig_topk_expert}. For a deeper analysis, we explore how different initialization methods affect expert selection. As shown in Fig.~\ref{fig_apdx_topk_expert}, both random initialization and co-upcycle result in a more uniform distribution of experts. However, this uniformity does not inherently lead to improved performance or interpretability, possibly due to insufficient differentiation among the experts. In contrast, our \name\ clearly prefers specialized roles, as illustrated in Fig.~\ref{fig_token_wise_expert}\&~\ref{fig_apdx_tokenwise_llava}\&~\ref{fig_apdx_tokenwise_chart}.

\subsection{Summary of hyperparameter settings}
\label{apdx_hyper_params}
The training process of our \name\ is structured into three distinct phases: Alignment Pre-training, High-Quality Knowledge Learning, and Chart-Specific Annealing Tuning. Table \ref{tab_hparam_appendix} provides a comprehensive overview of the hyperparameter configurations employed during each training stage. It should be noted that using flash-attention~\cite{flashattention2} can effectively reduce GPU hours. 

\begin{table*}[h]
\centering
\captionsetup{skip=0pt}
\caption{Training hyperparameters of \name \ for all stages.}
\renewcommand{\arraystretch}{1.5}
\resizebox{\linewidth}{!}{
\setlength{\tabcolsep}{2pt}
\begin{tabular}{lccc}
\toprule[1.5pt]
\textbf{Configuration}           & \textbf{Alignment Pre-training} & \textbf{High-Quality Knowledge Learning} & \textbf{Chart Specific Annealing Tuning} \\ \hline
Connector Initialization                        & InternlmXC-v2        & Table\&JSON\&Code Experts + InternlmXC-v2                & \name\ 2nd-stage               \\ 
LLM Training                        & Freeze               & LoRA                         & LoRA               \\ 
Image Resolution              &   490                 & 490                     &     490                         \\ 
ViT Sequence Length               &    1225               & 1225                     &   1225                          \\ 
Optimizer                        &         AdamW          & AdamW                     &           AdamW                  \\ 
Optimizer Hyperparameter         &            $\beta_1 = 0.9, \beta_2 = 0.95, \epsilon = 1e^{-8}$      & $\beta_1 = 0.9, \beta_2 = 0.95, \epsilon = 1e^{-8}$                     &       $\beta_1 = 0.9, \beta_2 = 0.95, \epsilon = 1e^{-8}$                  \\ 
Peak Learning Rate               & $5e^{-5}$             & $5e^{-5}$                        & $1e^{-5}$                       \\ 
Learning Rate Schedule           &    cosine decay                   & cosine decay                     &        cosine decay                         \\ 
Weight Decay                     &       0.1                & 0.1                     &              0.1                   \\ 
Gradient Clip                    &          1.0             & 1.0                     &               1.0                  \\ 
Warm-up Ratio                    &           0.01            & 0.01                     &                       0.01          \\ 
Global Batch Size                & 256                 & 64                             & 64                             \\ 
Gradient Acc.                    &        16               & 8                     &                8                 \\ 
Numerical Precision               &        bfloat16                & bfloat16                     &           bfloat16                       \\ 
Optimizer Sharding               &      \checkmark                 & \checkmark                       &           \checkmark                      \\ 
Gradient Sharding               &       \checkmark                & \checkmark                       &            \checkmark                     \\ 
Parameter Sharding               &       $\times$                & $\times$                     &       $\times$                          \\ 
Activation Checkpointing         &        \checkmark               & \checkmark                           &          \checkmark                       \\ 
GPU Hours (A100-40G)         &  210   &    100     &   56          \\
\bottomrule[1.5pt]
\end{tabular}}
\label{tab_hparam_appendix}
\end{table*}

\subsection{Reproducibility Statement}
\label{apdx_repro}
We have included the architecture of \name \ in Section \ref{sec_method_1} and the complete training procedure in Section \ref{sec_method_2} and Section \ref{sec_method_3}. The training data recipe is listed in Tab.~\ref{tab_dataset} in detail. Hyper-parameter settings are shown in Appendix \ref{apdx_hyper_params}. We also introduce the generation pipeline for \alignname \ in Section \ref{sec_method_2}, and some detailed examples in Appendix \ref{apdx_sec_chartmoealign}. Furthermore,
our \alignname \ dataset and checkpoints of \name \ will be released soon on GitHub and Huggingface.

\clearpage

\section{Additional Vsualizations of Top-1 Expert Selection}
\label{apdx_sec_visual_token}
\begin{figure}[h]
    \begin{subfigure}[b]{0.25\textwidth}
        \includegraphics[width=\textwidth]{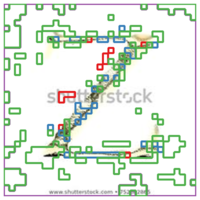}
    \end{subfigure}%
    \begin{subfigure}[b]{0.25\textwidth}
        \includegraphics[width=\textwidth]{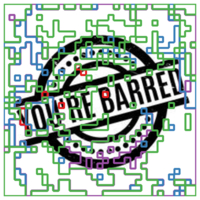}
    \end{subfigure}%
    \begin{subfigure}[b]{0.25\textwidth}
        \includegraphics[width=\textwidth]{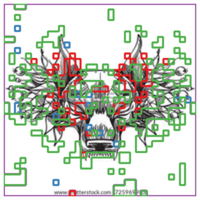}
    \end{subfigure}%
    \begin{subfigure}[b]{0.25\textwidth}
        \includegraphics[width=\textwidth]{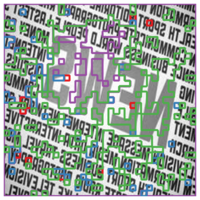}
    \end{subfigure}%
    \par
    \begin{subfigure}[b]{0.25\textwidth}
        \includegraphics[width=\textwidth]{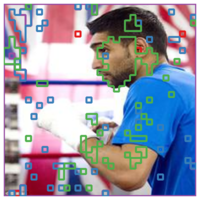}
    \end{subfigure}%
    \begin{subfigure}[b]{0.25\textwidth}
        \includegraphics[width=\textwidth]{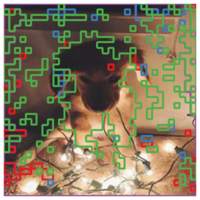}
    \end{subfigure}%
    \begin{subfigure}[b]{0.25\textwidth}
        \includegraphics[width=\textwidth]{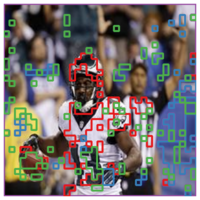}
    \end{subfigure}%
    \begin{subfigure}[b]{0.25\textwidth}
        \includegraphics[width=\textwidth]{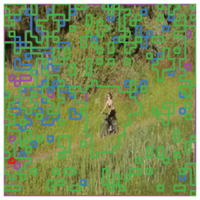}
    \end{subfigure}%
    \par
    \begin{subfigure}[b]{0.25\textwidth}
        \includegraphics[width=\textwidth]{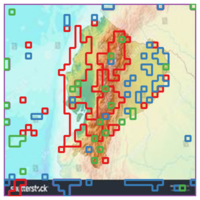}
    \end{subfigure}%
    \begin{subfigure}[b]{0.25\textwidth}
        \includegraphics[width=\textwidth]{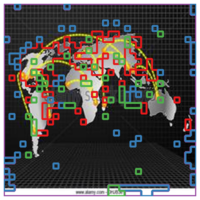}
    \end{subfigure}%
    \begin{subfigure}[b]{0.25\textwidth}
        \includegraphics[width=\textwidth]{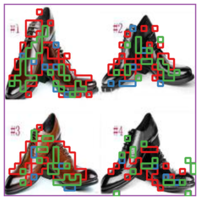}
    \end{subfigure}%
    \begin{subfigure}[b]{0.25\textwidth}
        \includegraphics[width=\textwidth]{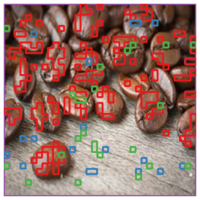}
    \end{subfigure}%
    \par
    \begin{subfigure}[b]{0.25\textwidth}
        \includegraphics[width=\textwidth]{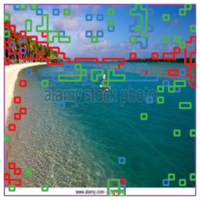}
    \end{subfigure}%
    \begin{subfigure}[b]{0.25\textwidth}
        \includegraphics[width=\textwidth]{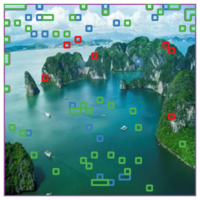}
    \end{subfigure}%
    \begin{subfigure}[b]{0.25\textwidth}
        \includegraphics[width=\textwidth]{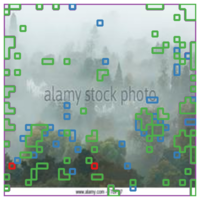}
    \end{subfigure}%
    \begin{subfigure}[b]{0.25\textwidth}
        \includegraphics[width=\textwidth]{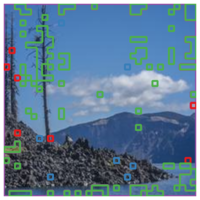}
    \end{subfigure}%
    \par
    \begin{subfigure}[b]{0.25\textwidth}
        \includegraphics[width=\textwidth]{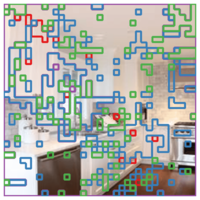}
    \end{subfigure}%
    \begin{subfigure}[b]{0.25\textwidth}
        \includegraphics[width=\textwidth]{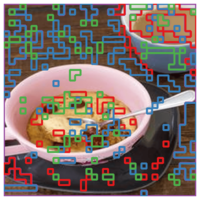}
    \end{subfigure}%
    \begin{subfigure}[b]{0.25\textwidth}
        \includegraphics[width=\textwidth]{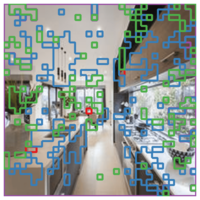}
    \end{subfigure}%
    \begin{subfigure}[b]{0.25\textwidth}
        \includegraphics[width=\textwidth]{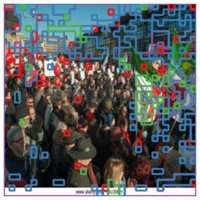}
    \end{subfigure}%
    \par
    \includegraphics[width=\textwidth]{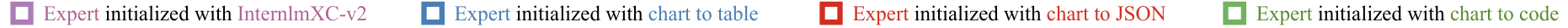}
    \centering
    \captionsetup{skip=5pt}
    \caption{More visualizations of top-1 expert selection on \textbf{\textit{general images}} randomly sampled from LLaVA-CC3M. These examples show the selection preferences of different experts in \name.}
    \label{fig_apdx_tokenwise_llava}
\end{figure}

\begin{figure}[t]
    \begin{subfigure}[b]{0.25\textwidth}
        \includegraphics[width=\textwidth]{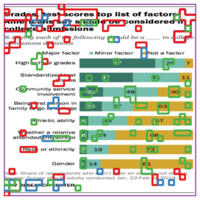}
    \end{subfigure}%
    \begin{subfigure}[b]{0.25\textwidth}
        \includegraphics[width=\textwidth]{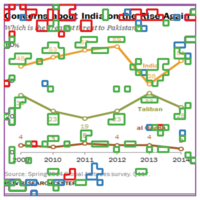}
    \end{subfigure}%
    \begin{subfigure}[b]{0.25\textwidth}
        \includegraphics[width=\textwidth]{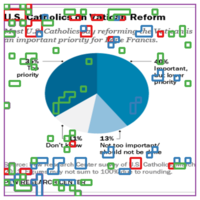}
    \end{subfigure}%
    \begin{subfigure}[b]{0.25\textwidth}
        \includegraphics[width=\textwidth]{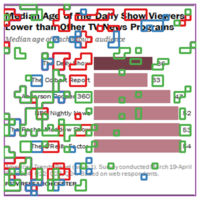}
    \end{subfigure}%
    \par
    \begin{subfigure}[b]{0.25\textwidth}
        \includegraphics[width=\textwidth]{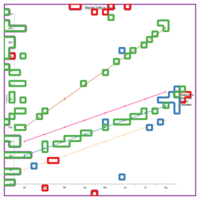}
    \end{subfigure}%
    \begin{subfigure}[b]{0.25\textwidth}
        \includegraphics[width=\textwidth]{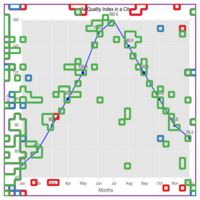}
    \end{subfigure}%
    \begin{subfigure}[b]{0.25\textwidth}
        \includegraphics[width=\textwidth]{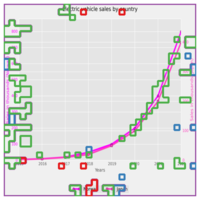}
    \end{subfigure}%
    \begin{subfigure}[b]{0.25\textwidth}
        \includegraphics[width=\textwidth]{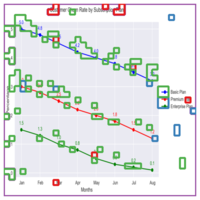}
    \end{subfigure}%
    \par
    \begin{subfigure}[b]{0.25\textwidth}
        \includegraphics[width=\textwidth]{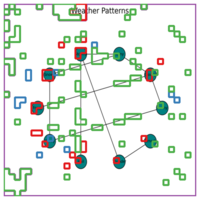}
    \end{subfigure}%
    \begin{subfigure}[b]{0.25\textwidth}
        \includegraphics[width=\textwidth]{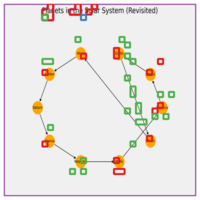}
    \end{subfigure}%
    \begin{subfigure}[b]{0.25\textwidth}
        \includegraphics[width=\textwidth]{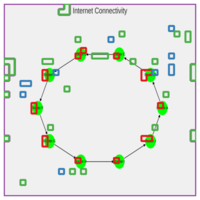}
    \end{subfigure}%
    \begin{subfigure}[b]{0.25\textwidth}
        \includegraphics[width=\textwidth]{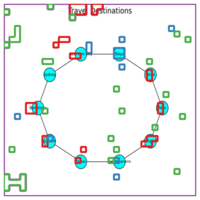}
    \end{subfigure}%
    \par
    \begin{subfigure}[b]{0.25\textwidth}
        \includegraphics[width=\textwidth]{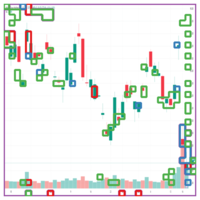}
    \end{subfigure}%
    \begin{subfigure}[b]{0.25\textwidth}
        \includegraphics[width=\textwidth]{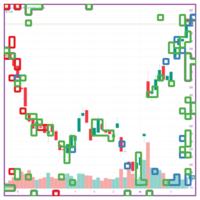}
    \end{subfigure}%
    \begin{subfigure}[b]{0.25\textwidth}
        \includegraphics[width=\textwidth]{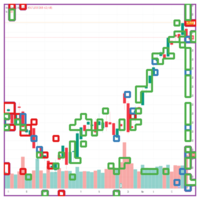}
    \end{subfigure}%
    \begin{subfigure}[b]{0.25\textwidth}
        \includegraphics[width=\textwidth]{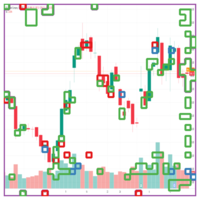}
    \end{subfigure}%
    \par
    \begin{subfigure}[b]{0.25\textwidth}
        \includegraphics[width=\textwidth]{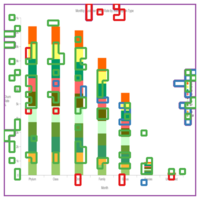}
    \end{subfigure}%
    \begin{subfigure}[b]{0.25\textwidth}
        \includegraphics[width=\textwidth]{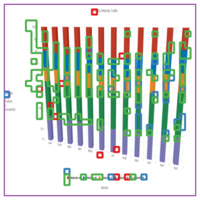}
    \end{subfigure}%
    \begin{subfigure}[b]{0.25\textwidth}
        \includegraphics[width=\textwidth]{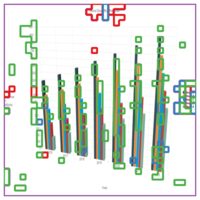}
    \end{subfigure}%
    \begin{subfigure}[b]{0.25\textwidth}
        \includegraphics[width=\textwidth]{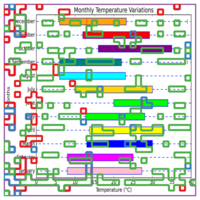}
    \end{subfigure}%
    \par
    \begin{subfigure}[b]{0.25\textwidth}
        \includegraphics[width=\textwidth]{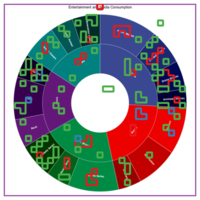}
    \end{subfigure}%
    \begin{subfigure}[b]{0.25\textwidth}
        \includegraphics[width=\textwidth]{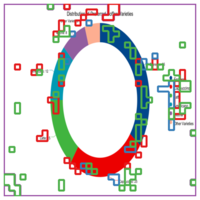}
    \end{subfigure}%
    \begin{subfigure}[b]{0.25\textwidth}
        \includegraphics[width=\textwidth]{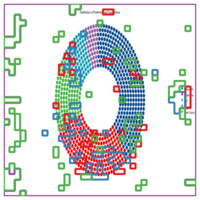}
    \end{subfigure}%
    \begin{subfigure}[b]{0.25\textwidth}
        \includegraphics[width=\textwidth]{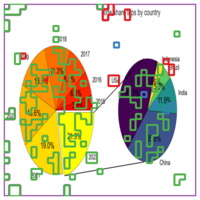}
    \end{subfigure}%
    \par
    \includegraphics[width=\textwidth]{Fig/figs/appendix_tokenwise_chart/tokenwise_legend.pdf}
    \centering
    \captionsetup{skip=3pt}
    \caption{More visualizations of top-1 expert selection on \textbf{\textit{chart images}}. The vanilla expert primarily handles background tokens, and the chart visual markers are handled by other experts.}
    \label{fig_apdx_tokenwise_chart}
\end{figure}

In this section, we randomly sampled images from natural image datasets (LLaVA-CC3M~\cite{llava}) and chart datasets (ChartQA~\cite{ChartQA}, ChartBench~\cite{ChartBench}) to illustrate ChartMoE's token selection preferences. As shown in Fig.~\ref{fig_apdx_tokenwise_llava}, the vanilla expert focuses more on the background, the table expert concentrates on details such as the boundary between the background and the subject, the JSON expert focuses on textures (e.g., maps and objects), and the code expert specializes in curves and trends (e.g., logos and text). Fig.~\ref{fig_apdx_tokenwise_chart} further demonstrates that while the vanilla expert continues to attend to background tokens, critical visual elements are handled by the aligned experts, with the code expert being notably more prominent.

\clearpage
\section{Details of \alignname}
\label{apdx_sec_chartmoealign}

\subsection{Overview}
\alignname\ is a dataset we introduced for different experts aligning pretraining. It consists of nearly 1 million Chart Table JSON Code quadruples and supports three alignment tasks: Chart to Table, Chart to JSON, and Chart to Code. Unlike other chart datasets, \alignname\ focuses solely on these three fundamental alignment tasks without considering the diversity of instruction tasks.
\vspace{-5pt}

\subsection{Table Data Collection}
We primarily collect table data from three sources: the ChartQA training set~\cite{ChartQA}, the PlotQA training set~\cite{PlotQA}, and ChartY provided by OneChart~\cite{OneChart}.

\textbf{ChartQA} includes 18.3K real-world charts and provides accompanying meta tables. While the charts are of high quality and manually curated, they lack fine-grained attribute annotations and executable plotting code. As a result, we only retained the tables from ChartQA in CSV format.

\textbf{PlotQA} comprises 157K charts, primarily focusing on three common types: line, bar, and pie charts. These charts are generated using Python code with limited formatting and style diversity. Consequently, we did not utilize the charts from PlotQA but retained its 157K tables. These tables originate from sources like World Bank Open Data, Open Government Data, and the Global Terrorism Database, covering statistics on various indicators such as fertility rates, rainfall, coal production, and more across years, countries, and districts.

\textbf{ChartY} is a chart dataset containing 2.7M charts in both Chinese and English proposed by OneChart. Notably, ChartY also includes charts from ChartQA and PlotQA, which we filtered out in \alignname. Additionally, ChartY primarily consists of common chart types such as line, bar, and pie charts (or their combinations) and suffers from significant data imbalance. To address this, we sampled a subset to ensure a roughly equal number of charts for each type. As the tables in ChartY are mainly generated by GPT-3.5 based on templates, we ultimately retained 763K samples from this source.
\vspace{-5pt}

\subsection{Pair Data Construction}
\textbf{JSON} provides a structured format distinct from CSV, designed to retain chart attributes beyond numerical data, such as title position, font size, element colors, legend styles, and more. We adopt the template provided by ChartReformer~\cite{ChartReformer} and further enhance it. We add chart type-agnostic attributes like title position and gridlines. For chart type-specific attributes, we aim to remain consistent with ChartReformer's definitions while accommodating all chart types present in \alignname. With this framework, we generate corresponding JSON files for all tables. To extract chart type-specific attributes, we fine-tune a Deplot~\cite{DePlot} model, leveraging the original chart to extract their properties. Missing attributes are filled in using random sampling to ensure completeness.

\textbf{Code} refers to Python scripts based on \textit{matplotlib} for rendering the charts. Leveraging the rich attributes defined in the JSON, the code is designed to faithfully represent every attribute to ensure diversity in the resulting charts. During generation, we explicitly specify all default parameters, such as the hexadecimal color codes for each line/bar, default font sizes, text positions, etc. We provide basic code templates for type-agnostic attributes. For type-specific attributes, rules are used to automatically generate the corresponding code.

\textbf{Chart} is produced by executing the generated code. Given the number of table, JSON, and code pairs, we filter out any quadruples with execution errors or warnings during the chart generation process, retaining only valid and error-free samples.

\textbf{Instruction}. Considering the alignment task, we directly employ several templated questions to define the Chart-to-X tasks (X is the ground truth). Ultimately, each quadruple corresponds to three QA pairs. Note that \alignname\ only serves for alignment training to initialize different expert projectors, thus emphasizing the diversity of charts and aligned modalities. To improve model performance and instruction-following, we still require more diverse instructions for supervised fine-tuning to update the MoE connector and LLM.
\vspace{-5pt}

\subsection{Quality Control}
We first remove all duplicate entries from the meta table and then eliminate quadruples that cause errors or warnings during rendering. To further assess the quality of \alignname, we randomly sample 200 quadruples and ask GPT-4o and annotation experts (with at least three experts reviewing each quadruple) to evaluate the clarity and readability of the charts, as well as the alignment between the charts and table/JSON/code, scoring them as 0 or 1. The results show that nearly all charts are clear, unambiguous, and free from obstructions (GPT-4o: 96.5\%, Experts: 99\%). Over 90\% of the pairs are matching and suitable for instruction tuning (GPT-4o: 91\%, Experts: 94.5\%).

\clearpage
\subsection{Example Visualization}
\begin{figure}[h]
    \begin{subfigure}[b]{\textwidth}
        \includegraphics[width=\textwidth]{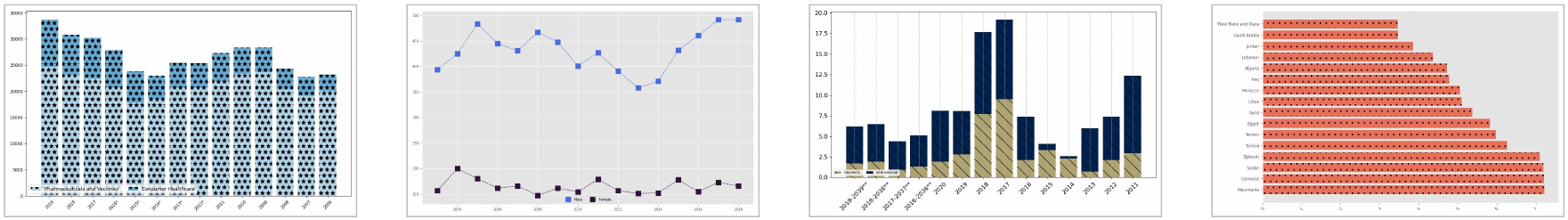}
        \captionsetup{skip=5pt}
        \caption{Charts in \alignname.}
    \end{subfigure}%\
    \par\vspace{5pt}
    \begin{subfigure}[b]{\textwidth}
        \includegraphics[width=\textwidth]{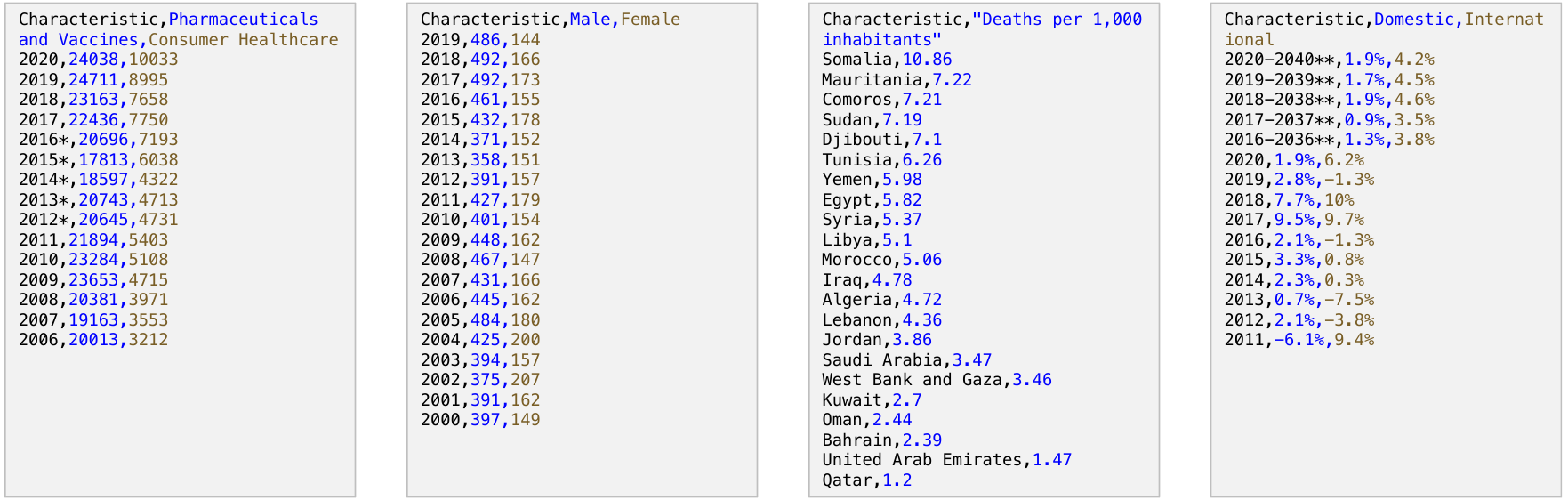}
        \captionsetup{skip=5pt}
        \caption{Tables in \alignname.}
    \end{subfigure}%
    \par\vspace{5pt}
    \begin{subfigure}[b]{\textwidth}
        \includegraphics[width=\textwidth]{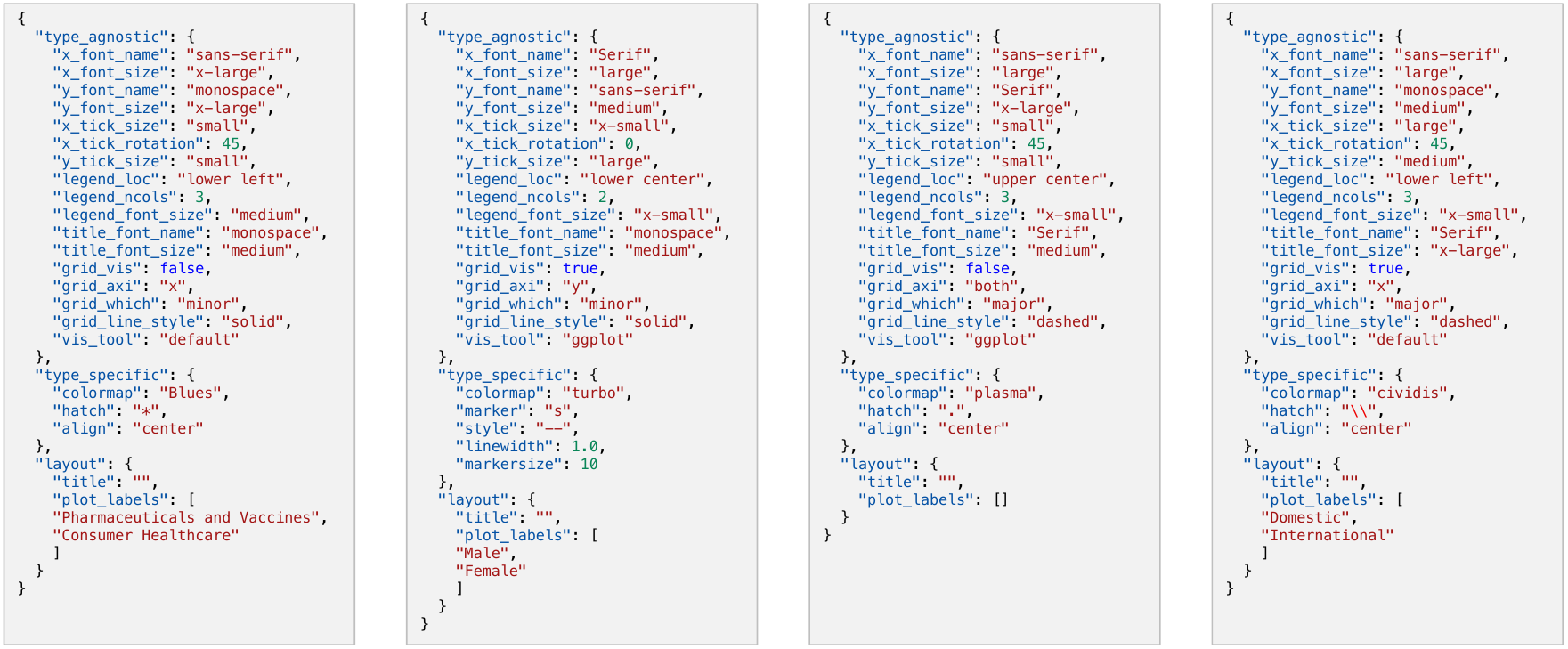}
        \captionsetup{skip=5pt}
        \caption{JSONs in \alignname. JSON is combined with the table during alignment pre-training.}
    \end{subfigure}%
    \par\vspace{5pt}
    \begin{subfigure}[b]{\textwidth}
        \includegraphics[width=\textwidth]{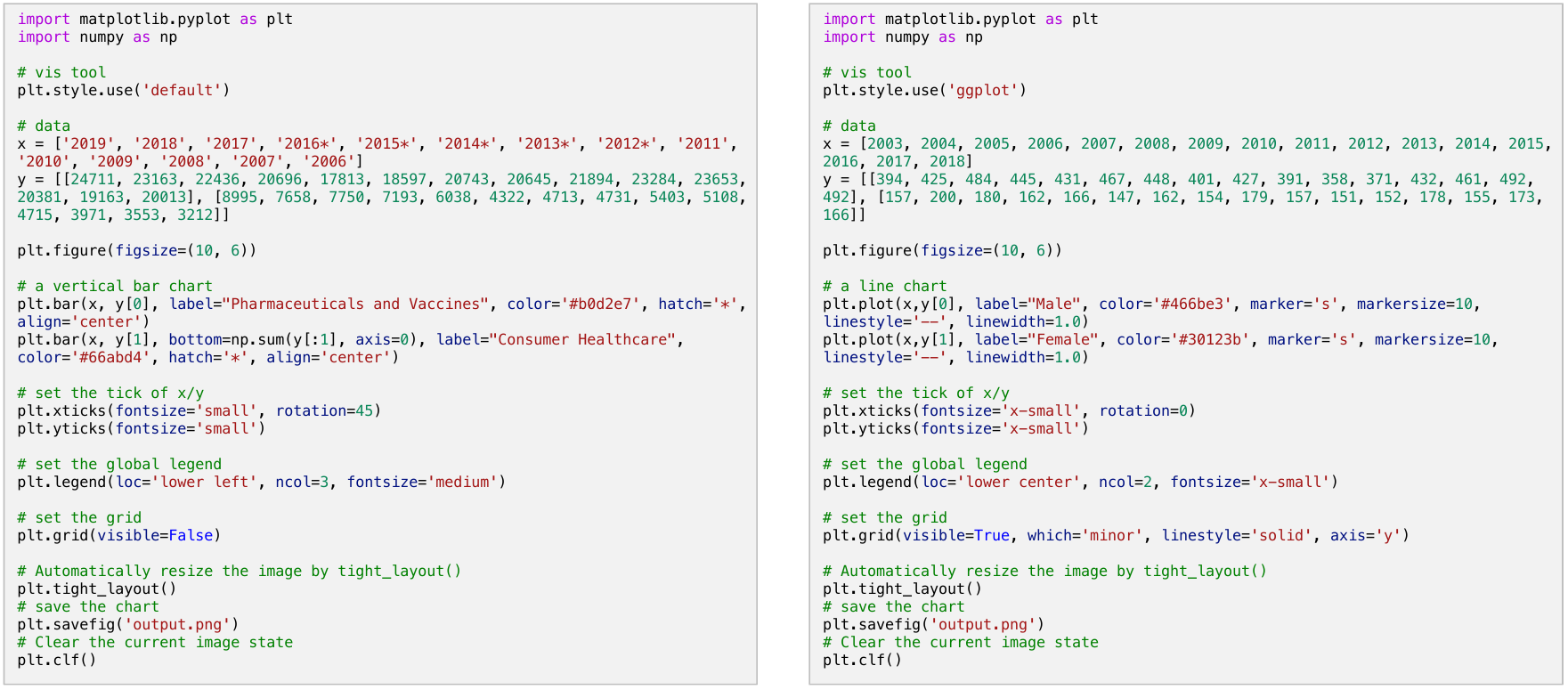}
        \captionsetup{skip=5pt}
        \caption{Codes in \alignname. All values and attributes are expressed explicitly.}
    \end{subfigure}%
    \centering
    \captionsetup{skip=3pt}
    \caption{Detailed Examples in \alignname. Each quadruple contains the chart, table, JSON and code.}
    \label{fig_example_of_chartalign}
\end{figure}

% \clearpage
% \section{Visually-grounded Code Generation Capabilities}
% \TODO{add details}
% \input{Tab/tab_chartmimic}

\clearpage
\section{Further Discussion}

\subsection{Contribution of \name}
\label{apdx_contribution}
Some prior work, such as MoE-LLaVA~\cite{moe-llava}, DeepSeek-VL~\cite{deepseek}, and CuMo~\cite{cumo}, has employed MoE architectures in MLLMs. However, these approaches all apply MoE to LLMs or ViTs to increase model capacity, introducing a large number of learnable parameters to boost performance. In contrast, our \name\ introduces several distinctive innovations:

\textbf{1) Motivation}: Our goal is not to expand model capacity but to enhance the model's chart comprehension through alignment tasks while preserving performance on other general tasks. Hence, we retain the original connector parameters as one expert initialization manner.

\textbf{2) Initialization}: Unlike previous methods that rely on random or co-upcycle initialization, we leverage multiple alignment tasks for expert (connector) initialization. This approach enables \name\ to exhibit remarkable interpretability (Fig.~\ref{fig_token_wise_expert}\&~\ref{fig_apdx_tokenwise_llava}\&~\ref{fig_apdx_tokenwise_chart}).

\textbf{3) Complexity}: We are the first to apply MoE exclusively to the MLP connector (projector) in LLaVA-like MLLMs. In \name\ (based on InternlmXC-v2), the MoE architecture introduces minimal additional parameters (model size 8.364B $\rightarrow$ 8.427B, + 63M$\uparrow$ only) and training complexity (Fig.~\ref{fig_loss}). It also shows negligible impact on inference speed (0.945 $\rightarrow$ 0.952 seconds per QA on ChartQA test set) and peak memory usage (23.72 GB $\rightarrow$ 23.86 GB, \textit{fp16} on A100-40G GPU).

\subsection{\name\ based on Other MLLMs}
Our \name\ is based on InterlmXC-v2, but our proposals (MoE connector, diverse alignment, etc.) are general approaches. Therefore, we use 10\% of the alignment data (Tab.~\ref{tab_dataset}) and the ChartQA training data to train our proposals based on LLaVA-v1.5-7B to further demonstrate their effectiveness. As shown in Tab.~\ref{tab_chartmoe_llava}, our proposals significantly improve the base model. This is partly because LLaVA is trained with fewer chart data, leading to a lower baseline, and also indicates that the additional alignment data greatly enhances chart understanding.

\begin{table}[h]
\centering
\captionsetup{skip=0pt}
\caption{Performance comparison on ChartQA with LLaVA-v1.5-7B as base MLLM.}
\resizebox{\linewidth}{!}{
\setlength{\tabcolsep}{12pt}
\begin{tabular}{@{}c|ccc|ccc|ccc@{}}
\toprule[1pt]
\multirow{2}{*}{Models} & \multicolumn{3}{c|}{Relax Acc @0.05} & \multicolumn{3}{c|}{Relax Acc @0.10} & \multicolumn{3}{c}{Relax Acc @0.20} \\ \cmidrule(l){2-10} 
 & Human & Aug & Avg & Human & Aug & Avg & Human & Aug & Avg \\ \midrule
LLaVA-v1.5-7B & 7.60 & 7.36 & 7.48 & 7.92 & 8.08 & 8.00 & 9.04 & 9.52 & 9.28 \\
LLaVA-v1.5-7B + ChartQA & 6.08 & 23.04 & 14.56 & 8.24 & 32.96 & 20.60 & 10.32 & 42.16 & 26.24 \\
LLaVA-v1.5-7B + ChartMoE & 18.13 & 32.11 & 25.12 & 20.20 & 42.32 & 31.36 & 24.24 & 52.12 & 38.18 \\ \bottomrule[1pt]
\end{tabular}
}
\label{tab_chartmoe_llava}
\end{table}

\subsection{Performance on ChartQA}
In Tab.~\ref{tab_chartqa}, our \name\ significantly outperforms SOTA. However, some models perform better than ours on the \textit{Augment} part of the ChartQA test set. Given that the \textit{Augment} part of ChartQA is considerably easier than the \textit{Human} part, we conduct a more detailed analysis. We analyze the performance of various models on numeric (\textit{Human}: 43\%, \textit{Augment}: 39\%) and non-numeric (\textit{Human}: 57\%, \textit{Augment}: 61\%) questions. As shown in Tab.~\ref{tab_chartqa_number}, \name\ excels in all subcategories except for non-numeric questions in the \textit{Augment} part. We find that \name's errors primarily occur in string-matching tasks. For instance, a prediction of \textit{It is between 2003 and 2005} is marked incorrect if the ground truth is \textit{(2003, 2005)}. High accuracy in this category may indicate overfitting instead.

\begin{table}[h]
\centering
\captionsetup{skip=0pt}
\caption{Fine-grained performance comparison on ChartQA with error margin 5\%.}
\resizebox{\linewidth}{!}{
\setlength{\tabcolsep}{12pt}
\begin{tabular}{@{}c|ccc|ccc|c@{}}
\toprule[1pt]
\multirow{2}{*}{Method} & \multicolumn{3}{c|}{Human} & \multicolumn{3}{c|}{Augment} & \multirow{2}{*}{Acc} \\ \cmidrule(lr){2-7}
 & Numeric & Non-Numeric & Avg & Numeric & Non-Numeric & Avg &  \\ \midrule
TinyChart & 58.52\% & 58.03\% & 58.24\% & 92.43\% & \textbf{96.25}\% & 94.32\% & 76.28\% \\
ChartAst & 67.04\% & 65.35\% & 66.08\% & 93.20\% & \textbf{93.07}\% & 93.12\% & 79.00\% \\
ChartMoE (Ours) & 73.89\% & 75.49\% & 74.80\% & 93.20\% & \textbf{90.98}\% & 91.84\% & 84.64\% \\ \bottomrule[1pt]
\end{tabular}
}
\label{tab_chartqa_number}
\end{table}

\subsection{Limitations}
\name \ has two limitations: 1) Dependency on alignment tasks. \name\ requires chart-Table/JSON/Code alignment tasks for initialization. Non-chart multimodal tasks need new alignment designs to initialize MoE experts. 2) Limited flexibility. Modifying the projector into a multi-expert architecture makes \name\ non-plug-and-play like LoRA. We are required to retrain the router network when new experts are coming.

\clearpage
\section{Capability Comparisons on Chart Editing and Highlighting}
\label{apdx_sec_chatdemo}
ChartMoE, leveraging carefully designed alignment training, excels in accurately extracting chart values and attributes, facilitating precise image reconstruction, chart element editing, and highlighting. In this section, we provide a visual comparison with the SOTA MLLMs, Qwen-2.5~\cite{Qwen25}, and GPT-4O~\cite{gpt4o}. While all three models perform the required instructions effectively, ChartMoE demonstrates superior accuracy in value and attribute extraction.

\begin{figure*}[h!]
    % \hsize=\textwidth
    \centering
      \begin{overpic}[width=0.9\linewidth, grid=False]{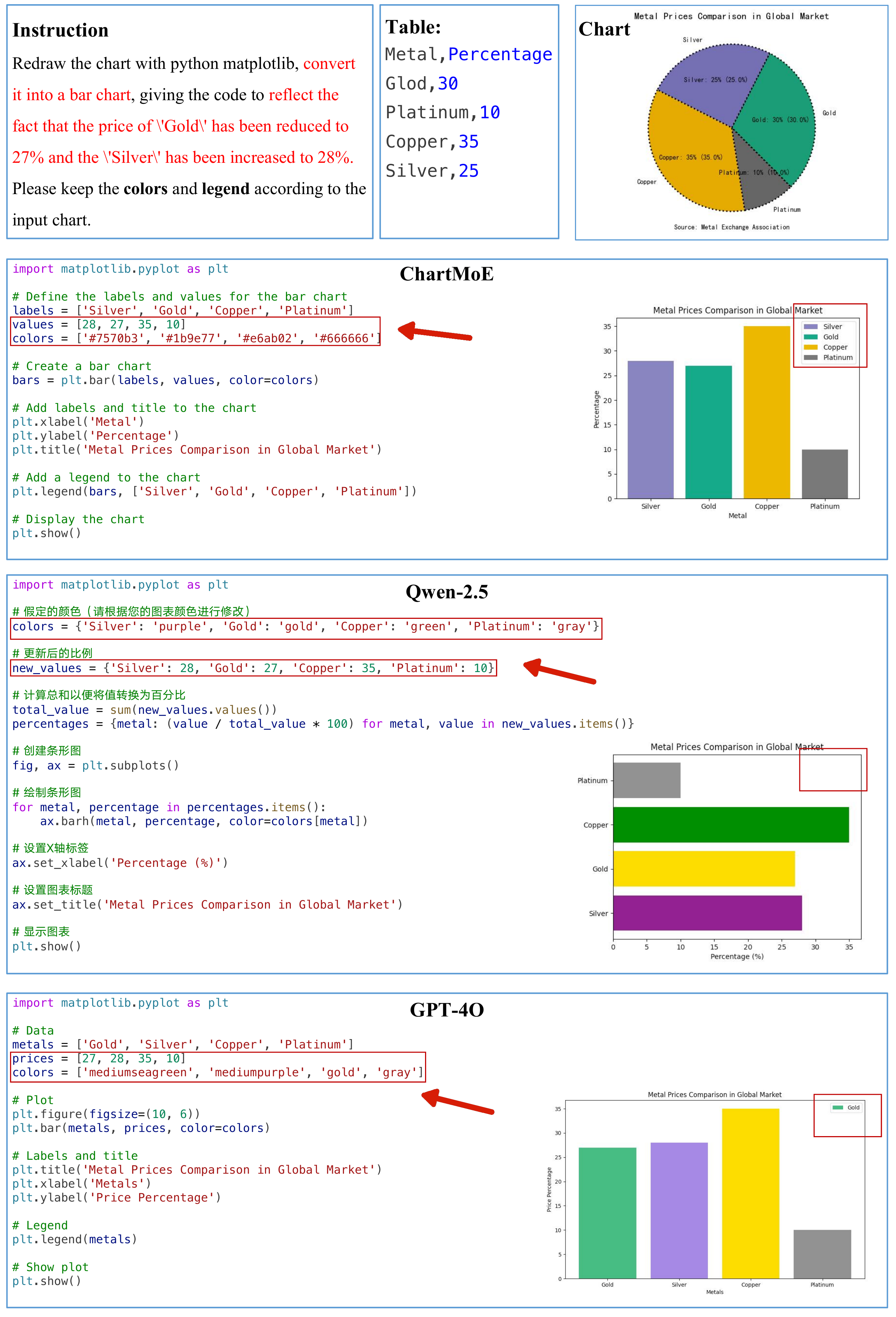}
      \end{overpic}
      \captionsetup{skip=0pt}      
      \caption{Chat demo involves \textbf{\textit{modifying chart types and values}}. All models successfully convert the chart type, but only ChartMoE handles the legend correctly.  No model makes errors in this task due to the simplicity of the values and the presence of data point labels.}
      % \vspace{-10pt}
    \label{fig_appendix_edit-demo1}
\end{figure*}

\clearpage
\begin{figure*}[t]
    % \hsize=\textwidth
    \centering
      \begin{overpic}[width=0.9\linewidth, grid=False]{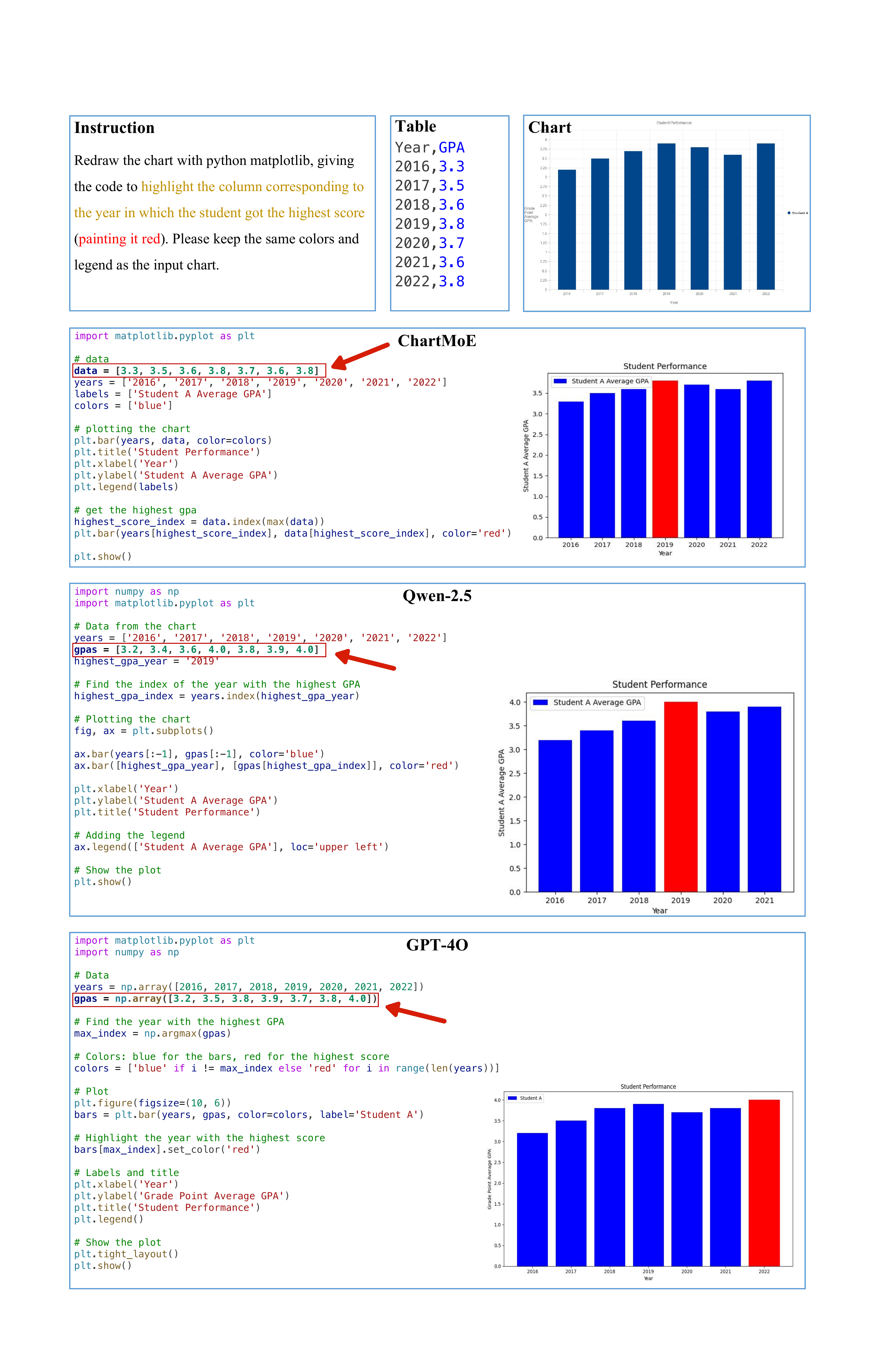}
      \end{overpic}
      \captionsetup{skip=0pt}      
      \caption{Chat demo involves modifying \textbf{\textit{chart editing}}. The bar chart is without labeled data points, and all methods provide reasonable and executable drawing code. Qwen-2.5 directly identifies the highest element, while the other two methods make it by code. Note that \name\ delivers the most accurate values, thanks to extensive alignment training and proposed MoE architecture.}
      % \vspace{-10pt}
    \label{fig_appendix_edit-demo2}
\end{figure*}

\clearpage
\begin{figure*}[t]
    % \hsize=\textwidth
    \centering
      \begin{overpic}[width=0.9\linewidth, grid=False]{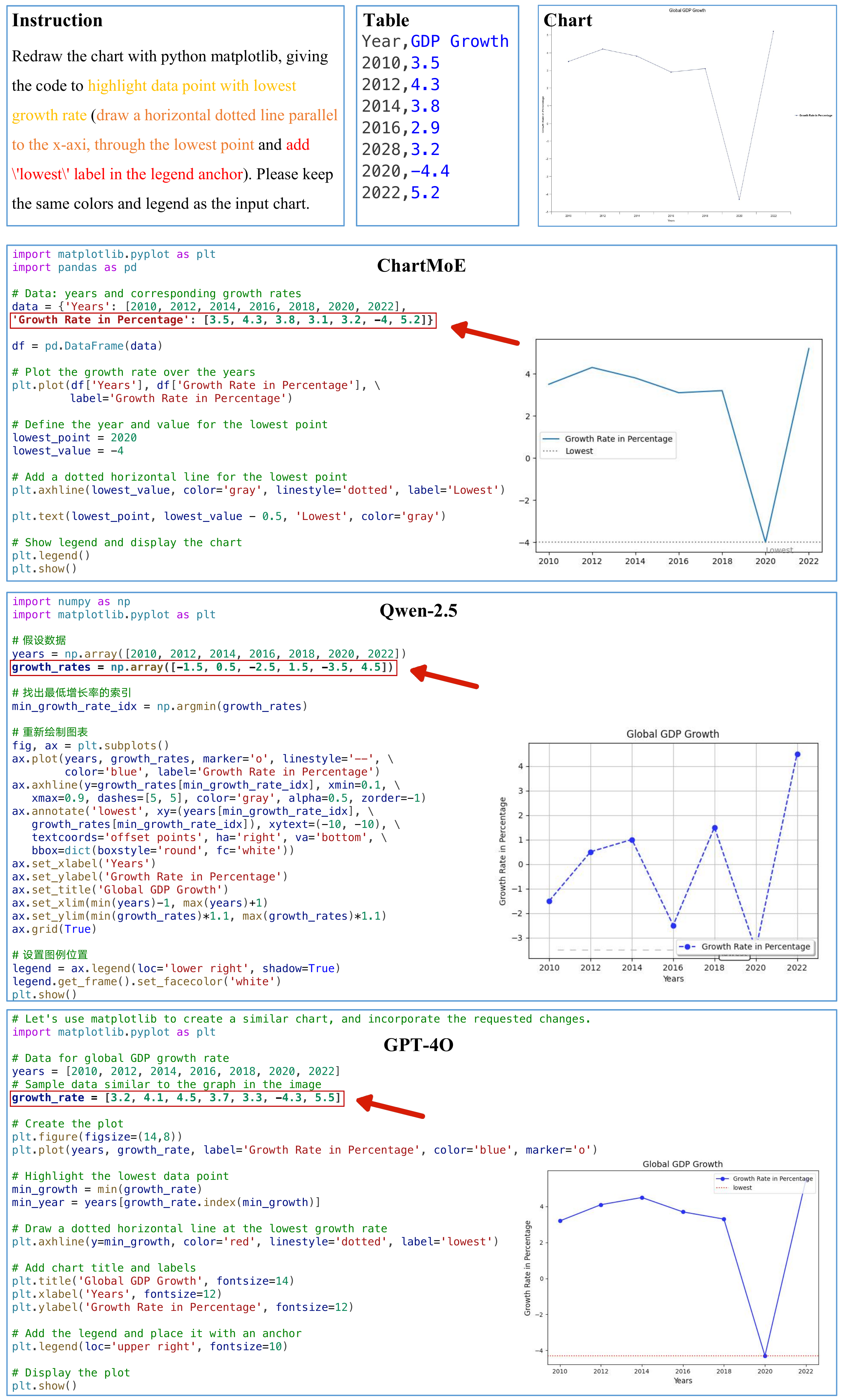}
      \end{overpic}
      \captionsetup{skip=5pt}      
      \caption{Chat demo involves modifying \textbf{\textit{chart editing}}. The line chart is without labeled data points, and all methods provide reasonable and executable drawing code. The values extracted by all models differ from the ground truth, but both ChartMoE and GPT-4O captured the correct data trends. Additionally, ChartMoE successfully completed all the editing tasks specified in the instructions.}
      % \vspace{-10pt}
    \label{fig_appendix_edit-demo3}
\end{figure*}

\end{document}